\documentclass[10pt,journal,compsoc]{IEEEtran}

\usepackage[nocompress]{cite}

\usepackage{hyperref} 
\usepackage{overpic}
\usepackage{graphicx,ragged2e}
\usepackage{amsmath}
\usepackage{amssymb}
\usepackage{array}
\usepackage{bm}
\usepackage{color}
\usepackage{pifont}
\usepackage[table]{xcolor}
\usepackage{algorithm}
\usepackage{algorithmic}
\usepackage{booktabs}
\usepackage{multirow}
\usepackage{graphbox} %loads graphicx package
\usepackage{url}
\usepackage{makecell}

\usepackage{silence}
\graphicspath{{./figure/}}

\newcolumntype{?}[1]{!{\vrule width #1}}
\newcolumntype{C}[1]{>{\centering\arraybackslash}p{#1}}

\newcommand{\abbest}[1]{{\bf {#1}}}
\newcommand{\best}[1]{{\color{red}{\bf{#1}}}}

\newcommand{\second}[1]{{\color{blue}{\underline{#1}}}}

\let\algref\cref
\newcommand{\figref}[1]{Fig.~\ref{#1}}
\newcommand{\tabref}[1]{Tab.~\ref{#1}}
\newcommand{\eqnref}[1]{Eqn.~(\ref{#1})}
\newcommand{\secref}[1]{Sec.~\ref{#1}}
\newcommand{\algref}[1]{Algorithm~\ref{#1}}
\newcommand{\etCite}[1]{\etal~\cite{#1}}

\newcommand{\smallsec}[1]{\vspace{8pt}\noindent\textbf{#1}}

\newcommand{\framework}{LED}
\newcommand{\dataset}{MultiRAW}

\newcommand{\eg}{\textit{e.g., }}
\newcommand{\ie}{\textit{i.e., }}
\newcommand{\etal}{\textit{et al. }}
\newcommand{\etc}{\textit{etc.}}

\definecolor{mygreen}{RGB}{112, 173, 71}
\definecolor{solution_area}{RGB}{68,114,196}

\newcommand{\twRow}[1]{\multirow{2}{*}{#1}}
\newcommand{\twCol}[1]{\multicolumn{2}{c}{#1}}

\newcommand{\tablestyle}[2]{\setlength{\tabcolsep}{#1}\renewcommand{\arraystretch}{#2}\centering}

\begin{document}
\title{Make Explicit Calibration Implicit:\\ 
Calibrate Denoiser Instead of the Noise Model}

\newcommand{\member}{\textit{~Member, IEEE}}
\newcommand{\smember}{\textit{~Senior Member, IEEE}}

\author{Xin~Jin, Jia-Wen~Xiao, Ling-Hao~Han, Chunle~Guo, Xialei Liu, Chongyi Li, and Ming-Ming Cheng
\IEEEcompsocitemizethanks{
\IEEEcompsocthanksitem All the authors are with VCIP, CS, Nankai University, Tianjin, China. 
% Email: xjin@mail.nankai.edu.cn
\IEEEcompsocthanksitem CL Guo and MM Cheng 
(\{guochunle,cmm\}@nankai.edu.cn) are corresponding authors.
\IEEEcompsocthanksitem This paper is an extension of our ICCV 2023 conference version~\cite{jin2023lighting}.
}% <-this % stops a space 
}

\IEEEtitleabstractindextext{%
\begin{abstract} \justifying
Explicit calibration-based methods have dominated RAW image denoising under extremely low-light environments. 
However, these methods are impeded by several critical limitations: a) the explicit calibration process is both labor- and time-intensive, b) challenge exists in transferring denoisers across different camera models, and c) the disparity between synthetic and real noise is exacerbated by digital gain. 
To address these issues, we introduce a groundbreaking pipeline named {\bf{L}}ighting {\bf{E}}very {\bf{D}}arkness ({\bf{\framework}}), which is effective regardless of the digital gain or the camera sensor.
\framework~eliminates the need for explicit noise model calibration, instead utilizing an implicit fine-tuning process that allows quick deployment and requires minimal data.
Structural modifications are also included to reduce the discrepancy between synthetic and real noise without extra computational demands.
Our method surpasses existing methods in various camera models, including new ones not in public datasets, with just a few pairs per digital gain and only 0.5$\%$ of the typical iterations.
Furthermore, \framework~also allows researchers to focus more on deep learning advancements while still utilizing sensor engineering benefits.
Code and related materials can be found in~\url{https://srameo.github.io/projects/led-iccv23/}.
\end{abstract}

\begin{IEEEkeywords}
Extreme low-light imaging, few-shot learning, 
deep low-light image denoising, low-light denoising dataset.
\end{IEEEkeywords}
}

% Make the title area
\maketitle
\IEEEdisplaynontitleabstractindextext

\IEEEraisesectionheading{\section{Introduction}}
\label{sec:intro}

Noise, an inescapable topic for image capturing, has been systematically investigated in recent years~\cite{buades2005non,zhang2017beyond,ulyanov2018deep,lehtinen2018noise2noise,abdelhamed2018high,chen2018learning,wei2021physics}.
Compared to standard RGB images, RAW images offer two substantial advantages for image denoising: tractable, primitive noise distribution~\cite{wei2021physics} and higher bit depth for differentiating signal from noise.
Learning-based methodologies have demonstrated remarkable advancements in RAW image denoising, particularly when utilizing paired real datasets~\cite{zhang2018ffdnet,guo2019toward,zamir2022learning,jin2023dnf}.
However, creating extensive real RAW image datasets tailored to each camera model is impractical.
Consequently, there has been a growing focus on applying learning-based techniques to synthetic datasets, a trend reflected in various studies~\cite{abdelhamed2019noise,zamir2020cycleisp,jang2021c2n,wei2021physics,zhang2021rethinking,maleky2022noise2noiseflow,kousha2022modeling}.

Calibration-based noise synthesis, particularly when employing physics-based models, has demonstrated its proficiency in accurately fitting real noise characteristics~\cite{wang2020practical,wei2021physics,zhang2021rethinking,monakhova2022dancing,zou2022estimating,feng2022learnability}.
These methods typically adhere to a systematic process.
Initially, they construct a well-designed noise model that aligns with the electronic imaging pipeline. 
Subsequently, a specific target camera is chosen, and the parameters of the pre-defined noise model are meticulously calibrated.
The final step involves generating synthetic paired data for training a denoising network.
Moreover, some approaches have been exploring the use of Deep Neural Network (DNN)-based generative models to facilitate the calibration of noise parameters~\cite{monakhova2022dancing,zou2022estimating}.

\begin{figure}[t]
  \centering
  \begin{overpic}[width=\linewidth]{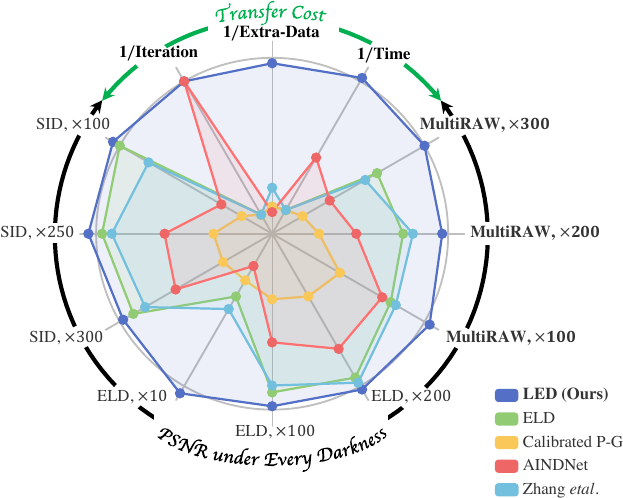}
    \put(89.0, 12.1){\scriptsize~\cite{wei2021physics}}
    \put(94.5, 4.6){\scriptsize~\cite{kim2020transfer}}
    \put(96.3, 0.8){\scriptsize~\cite{zhang2021rethinking}}
  \end{overpic}
  \vspace{-4mm}
  \caption{
    \framework~exhibits unparalleled state-of-the-art performance across a spectrum of darkness scenarios, encompassing various digital gain levels and camera sensors, outperforming calibration-based and transfer learning-based methodologies. 
    Furthermore, adopting our proposed pipeline for new camera models requires minimal cost. 
    {\textit{Metrics are scaled into non-linear space for best understanding.}}
    Refer to \secref{sec:exps} for a comprehensive explanation. 
  }\label{fig:teaser_comparison}
\end{figure}

\begin{figure}
  \centering
  \begin{overpic}[width=\linewidth]{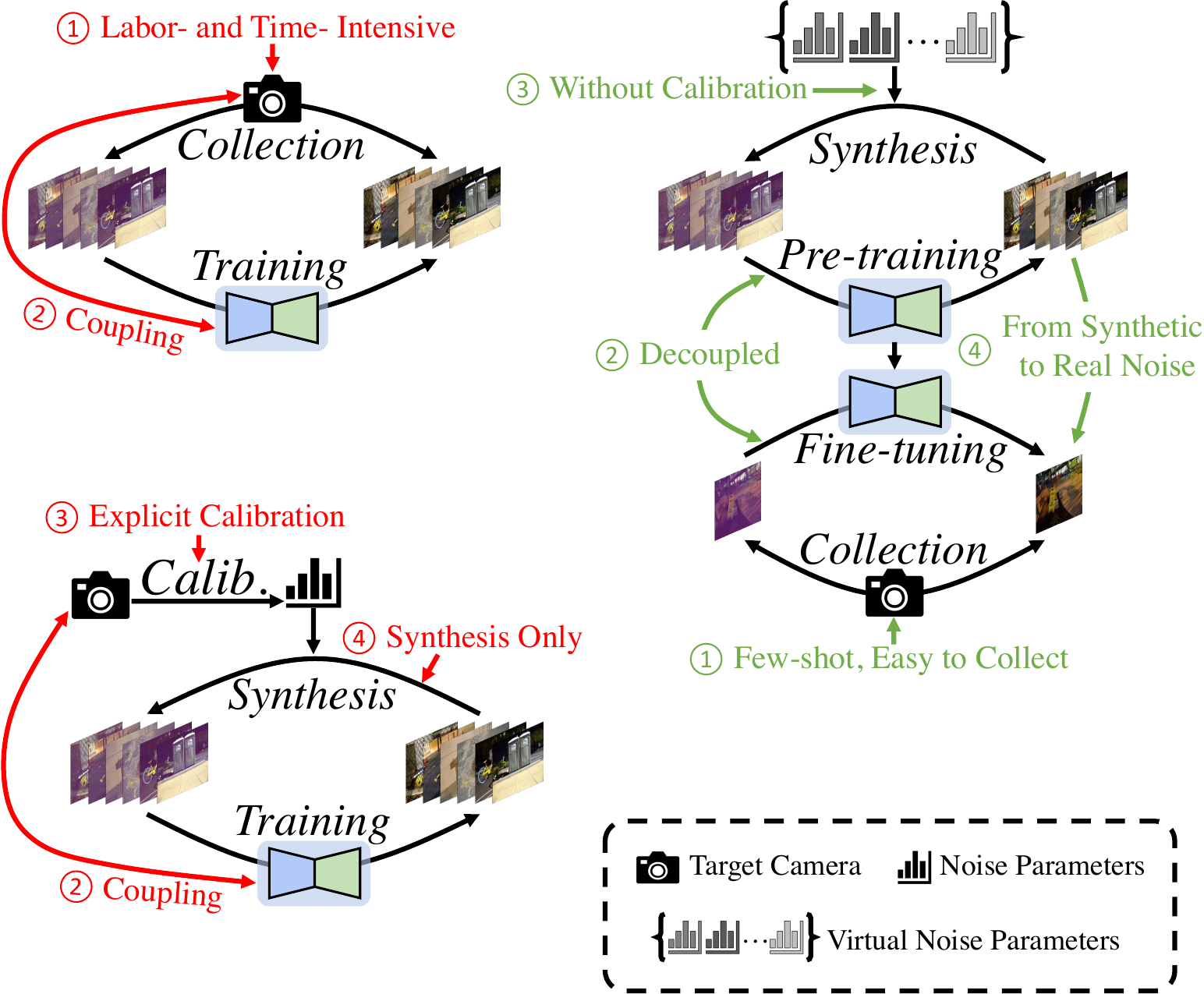}
    \put(0,48){\small{(a) Paired data-based methods}}
    \put(0,2){\small{(b) Calibration-based methods}}
    \put(54.9,22.5){\small{(c) Our proposed method}}
  \end{overpic}\vspace{-4mm}
  \caption{
    The thumbnail of paired data-based methods, explicit calibration-based methods, and our proposed \framework~({\it{Zoom-in for best view}}).
    The ``\textcolor{red}{$\rightarrow$}'' denotes the limitations of the paired data- and calibration-based methods,
    and the ``\textcolor{mygreen}{$\rightarrow$}'' highlights our solutions for the above limitations.
    {\it{Calib.}} represents the calibration operations, including pre-defining a noise model, 
    collecting calibration-specialized data, post-processing, and calculating the noise parameters.
    In \framework, the collection procedure only captures few-shot paired data, alleviating the deployment cost.
  }
  \label{fig:teaser_framework}
\end{figure}

Despite their notable achievements, current methods encounter three principal limitations, as depicted in \figref{fig:teaser_framework} (b).
1) Explicit camera-specific noisy model calibration is time-consuming and labor-intensive, requiring specialized data collection with a consistent illumination environment and comprehensive post-processing.
2) Each denoising network (denoiser) is tailored for a specific camera model.
Such coupling issues exhibit adaptability challenges to different cameras, requiring repeated calibration and training for distinct target cameras. 
3) The noise model trained with synthetic-only data may not encompass certain noise distributions, leading to what is termed as {\bf{out-of-model}} noise~\cite{wei2021physics,zhang2021rethinking,feng2022learnability}.
In other words, a domain gap persists between Synthetic Noise (SN) and Real Noise (RN).
While recent advancements~\cite{zou2022estimating} have concentrated on reducing calibration costs through DNN-based methods, issues related to the coupling of between networks and cameras, and out-of-model noise continue to increase training expenses and constrain overall performance. 

We introduce an innovative pipeline, \framework, for lighting every darkness, addressing the identified shortcomings of calibration-based methods. 
As illustrated in \figref{fig:teaser_framework} (c), our framework eliminates the necessity for calibration data and operations related to the noise model.
To sever the strong dependency between the denoising network and a specific target camera, we propose a dual-stage approach: pre-training with a virtual camera set\footnote{``Virtual'' cameras do not correspond to any real camera models but with reasonable noise parameters of the pre-defined noise model. It is sampled from a parameter space $\mathcal{S}$ with our proposed sampling strategy. Details can be found in \secref{sec:pretrain}.} 
followed by fine-tuning with few-shot pairs from a specific real camera. 
This strategy effectively decouples the network from being bound to a single camera model.
Concerning the disparity between a virtual and a target camera and the challenges posed by out-of-model noise, we introduce the Re-parameterized Noise Removal (RepNR) block. 
During the pre-training stage, the RepNR block has several camera-specific alignments (CSA).
Each CSA is responsible for learning the camera-specific information of a single virtual camera and aligning features to a shared space. 
Then, the common knowledge of {\bf{in-model}} (components that have been assumed as part of the noise model) noise is learned by a shared denoising convolution.
In the fine-tuning stage, we average all the CSAs of virtual cameras as initialization of the target camera.
Additionally, we integrate a parallel convolution branch for Out-of-Model Noise Removal (OMNR).
During the fine-tuning stage, \framework~implicitly ``calibrates'' the parameters of the denoiser, especially the CSAs, instead of explicitly calibrating the noise model.
Only 2 pairs for each ratio (additional digital gain) captured by the target camera, in a total of 6 raw image pairs, are used for learning to remove real noise (discussion on \textbf{why 2 pairs for each ratio} can be found in~\secref{sec:discuss}).
During deployment, all the RepNR blocks can be structurally parameterized~\cite{ding2021diverse,ding2021repvgg,chen2023instance} into a straightforward $3\times 3$ convolution without any extra computational cost, yielding a plain UNet~\cite{ronneberger2015u}. 

To comprehensively evaluate the efficacy of \framework~across diverse camera models, we introduce a novel dataset specifically tailored for {\textbf{Multi}}-camera and dark scene {\textbf{RAW}} image denoising, referred to as {\textbf{\dataset}}. 
This dataset is distinct in that it includes five different camera models that have never appeared before. 
A notable feature of \dataset~is its encompassment of various sensor sizes, ranging from full-frame cameras to APS-C format cameras, offering a more expansive and realistic testing ground. 
Furthermore, \dataset~dataset will be used in the CVPR 2024 and subsequent MIPI (Mobile Intelligent Photography \& Imaging) workshops. 
This utilization underscores its significance and potential impact in advancing the field of RAW image denoising, particularly in scenarios characterized by extremely low light conditions.

Compared to \framework, previous methods primarily focused on constructing noise models and calibrating noise parameters, namely sensor-related engineering. 
However, \framework~has focused on deep learning techniques like few-shot and transfer learning. 
Additionally, our method does not deviate from traditional noise modeling methods, which can still empower the pre-training stage of \framework.

Our principal contributions are concisely encapsulated as follows:
\begin{itemize}
\item We introduce a novel, implicit ``calibration'' pipeline for lighting every darkness, eliminating the need for additional calibration-related expenses for noise parameter calculation.
\item The implementation of Camera-Specific Alignments (CSA) mitigates the dependence of the denoising network on specific camera models. 
At the same time, the Out-of-Model Noise Removal (OMNR) mechanism facilitates few-shot transfer by learning the out-of-model noise of different sensors.
\item We release a new dataset, \dataset, encompassing various camera models, assorted scenes, and varying brightness levels. 
This dataset substantially enriches the current landscape of open-source datasets and addresses the prevalent limitation of limited camera variety.
\item Remarkably, our method requires only 2 RAW image pairs for each ratio and a mere 0.5$\%$ of the iterations typically needed by state-of-the-art methods (\figref{fig:teaser_comparison}).
\end{itemize}

Compared to the ICCV 2023~\cite{jin2023lighting} version, this journal extension includes several notable expansions. 
1) Experiments (\secref{sec:application_exp}) demonstrate that our method can be seamlessly integrated with {\textbf{various existing network architectures}} and explicit calibration methods, showcasing the broad applicability of our proposed pipeline. 
2) Furthermore, a discussion is provided on whether the network employs noise prior or image prior during denoising (detailed in \secref{sec:discuss}), serving as guidance for further research.
3) We provide a detailed process for few-shot dataset collection and considerations, laying the groundwork for widespread adoption of our implicit calibration pipeline,~\framework. 
4) Based on the remainder in 3), we introduce a new dataset, \dataset, featuring various camera models (not included in prior public datasets), multiple additional digital gains, and each setting encompassing two different ISO configurations. 
5) We plan to invigorate the RAW image denoising community by hosting a Few-shot RAW Image Denoising competition with the proposed \dataset~dataset at the CVPR 2024 workshop: Mobile Intelligent Photography \& Imaging.

\section{Related Work}
\label{sec:related_work}

The issue of image capture in extremely dark scenes has received widespread attention from numerous camera/smartphone manufacturers. 
This section will revisit denoising techniques such as training with paired data and methods based on noise model calibration.

\subsection{Training with Paired Real Data.}

The field of RAW data exploitation for image denoising has its roots in the groundbreaking work of the SIDD project~\cite{abdelhamed2018high}. 
Progress in this area has recently broadened to encompass traditional light image denoising and the more complex challenges inherent in extremely low-light conditions. 
This expansion is illustrated by notable studies such as SID~\cite{chen2018learning} and ELD~\cite{wei2021physics}.
While methodologies based on real noise have yielded encouraging results~\cite{zamir2020learning,chen2021hinet,zamir2021multi,zamir2022restormer,chen2022simple,zhang2023real}, their widespread application is hampered by the considerable effort required to compile extensive datasets of paired low and high-quality images.
To address this, employing training strategies that utilize paired low-quality raw images, exemplified by Noise2Noise~\cite{lehtinen2018noise2noise} and Noise2NoiseFlow~\cite{maleky2022noise2noiseflow}, offers an effective workaround to the tedious task of assembling noisy-clean image pairs. 
However, these techniques tend to under-perform in severe noise levels, especially in scenarios with extreme darkness~\cite{chen2018learning,wei2021physics}.

In this context, our \framework~aims to advance the understanding and effectiveness of real noise elimination. 
It incorporates insights from a limited number of paired images taken in extremely low-light conditions, thereby mitigating the data collection challenges associated with such environments.

\subsection{Calibration-Based Denoising.}

While alleviating the burden of compiling pairwise datasets, synthetic noise-based techniques encounter practical limitations.
Common noise models like Poisson and Gaussian significantly diverge from actual noise distributions in extremely low-light conditions~\cite{chen2018learning,wei2021physics}
\footnote{
Denoising under extremely low-light scenarios necessitates the application of additional digital gain (up to 300$\times$) to the input, thereby intensifying the domain gap between real and synthetic noise.
}.
In response, explicit calibration-based methods, simulating each noise component in electronic imaging pipelines~\cite{boie1992analysis,healey1994radiometric,gow2007comprehensive,irie2008technique,konnik2014high}, have thrived due to their reliability. 

ELD~\cite{wei2021physics} proposed a noise model that closely aligns with real noise characteristics, achieving notable performance in dark scenarios.
Zhang \etCite{zhang2021rethinking} acknowledged the complexity of modeling signal-independent noise sources and proposed a method that randomly samples such noise from dark frames. 
However, it still necessitates calibration for signal-dependent noise parameters (overall system gain).
Monakhova \etCite{monakhova2022dancing} devised a noise generator combining physics-based noise models with a generative adversarial framework~\cite{goodfellow2014generative}. 
Zou \etCite{zou2022estimating} pursued more accurate and concise calibration by employing contrastive learning~\cite{chen2020simple,he2020momentum} for parameter estimation.

Despite the impressive performance achieved by calibration-based methods, certain challenges persist. 
Stable illumination environments (\eg consistent brightness and temperature), calibration-specific data collection (\eg multiple images for each camera setting), and intricate post-processing tasks (\eg alignment, localization, and statistical analyses) are prerequisites for precisely estimating noise parameters.
Furthermore, repeated calibration and training processes are essential for distinct cameras, owing to the diversity of parameters and the nonuniform pre-defined noise model~\cite{wach2004noise,gow2007comprehensive,konnik2014high,maggioni2014joint}. Additionally, the domain gap between synthetic and real noise is not adequately addressed.

Our \framework~overcomes these challenges by replacing the explicit calibration procedure with implicitly calibrating the denoiser: a pre-training and fine-tuning framework and a RepNR block designed for noise removal, respectively.

\subsection{From Synthetic to Real Noise.}

The domain gap between real and synthetic noise, a fundamental challenge, becomes particularly pronounced when models trained on synthetic data are tested on real-world data. 
To bridge this gap, recent research has increasingly focused on employing techniques like Adaptive Instance Normalization (AdaIN)~\cite{huang2017arbitrary,karras2019style} and few-shot learning~\cite{hospedales2021meta,ye2022few,huang2022survey}, along with transfer learning~\cite{kim2020transfer} and domain adaptation~\cite{prabhakar2021few} strategies.
However, these approaches often struggle in extremely dark environments where the numerical instability caused by intense noise and high digital gain can impair signal reconstruction.

To address this, our framework introduces a novel camera-specific alignment strategy. 
This method reduces numerical instability and effectively separates camera-specific characteristics from the general attributes of the noise model. 
Moreover, unlike instance or layer normalization~\cite{ulyanov2016instance,ba2016layer}, our alignment operations can be reparameterized into a straightforward convolution, similar to custom batch normalization~\cite{ioffe2015batch}. 
This reparameterization ensures that our approach does not incur any additional computational burden.

\begin{figure*}[t]
  \centering
  \begin{overpic}[width=\textwidth]{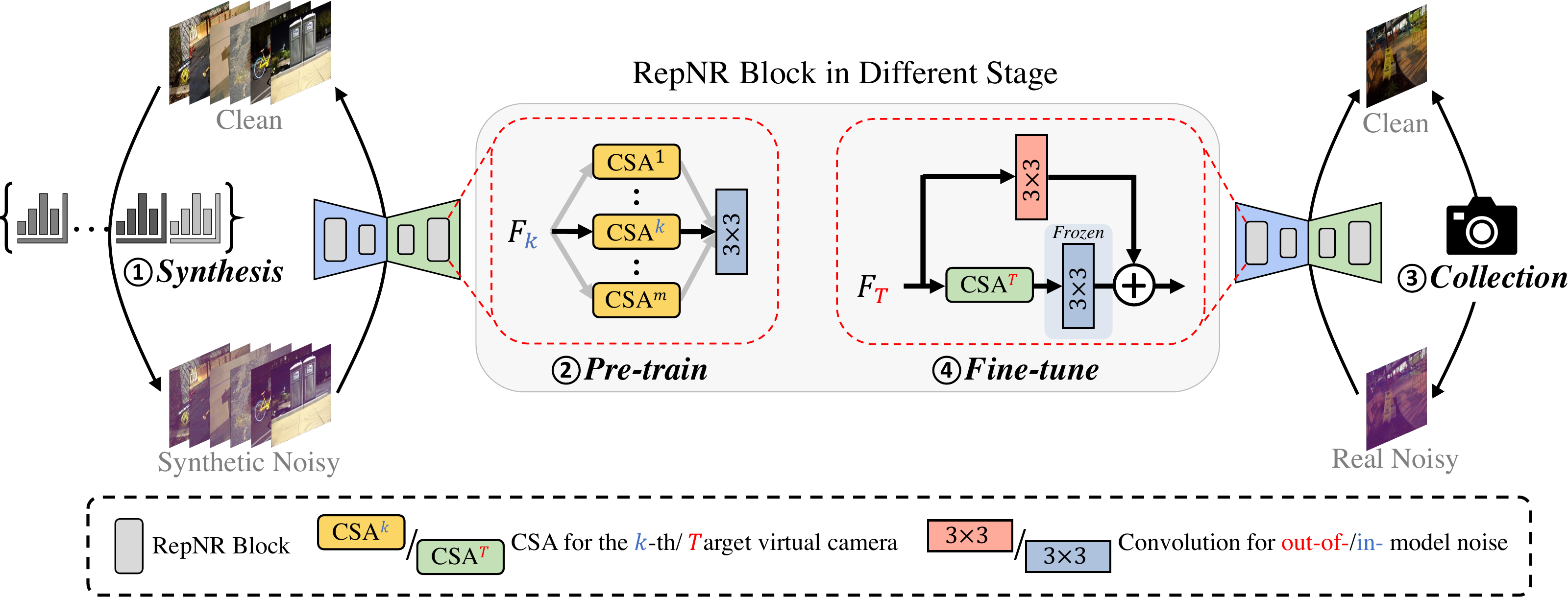}
    \put(49.9, 22.3){\LARGE\textcolor{mygreen}{$\Rightarrow$}}
  \end{overpic}
  \vspace{-2mm}
  \caption{
    Illustration of our proposed \framework~and RepNR block. The overall pipeline is delineated into four key stages:
    1) Sampling a set of $m$ virtual cameras responsible for synthesizing noise at a later stage;
    2) Pre-training the denoising network with $m$ camera-specific alignments (CSAs) and synthetic paired images, with each CSA corresponding to a virtual camera;
    3) Utilizing the target camera to acquire a limited number of real noisy image pairs;
    4) Fine-tuning the pre-trained denoising network with real noisy data, tailoring the network to the characteristics of the target camera.
    In the intermediary phase, we introduce distinct optimization strategies tailored for the specific training stages of our RepNR block.
    During the stage transition, indicated by ``{\textcolor{mygreen}{$\Rightarrow$}}'', we average the CSAs to initialize the CSA$^T$. Subsequently, once CSA$^T$ reaches convergence, we introduce the OMNR (\textcolor{red}{$3\times 3$}) branch alongside the existing IMNR (\textcolor{blue}{$3\times 3$} $+$ CSA$^T$) branch, and proceed with the training process.
  }
  \vspace{-3mm}
  \label{fig:archs}
\end{figure*}

\section{Method}
\label{sec:method}

This section commences with an overview of the complete pipeline for our proposed raw image denoising with implicit calibration. 
Subsequently, we introduce our Reparameterized Noise Removal (RepNR) block. 
The comprehensive denoising pipeline is illustrated in \figref{fig:archs}.

\subsection{Preliminaries and Motivation}

\def\TL{\mathcal{T}}

In raw image space, the captured signals $D$ are conventionally regarded as the sum of the clean image $I$ and various noise components $N$, expressed as \eqnref{eq:nc}.
\begin{equation}\label{eq:nc}
  D = I + N,
\end{equation}
where $N$ is assumed to follow a noise model,
\begin{align}\label{eq:nm}
    N = N_{shot} + N_{read} + N_{row} + N_{quant} + \epsilon,
\end{align}
with $N_{shot}$, $N_{read}$, $N_{row}$, $N_{quant}$, and $\epsilon$ representing shot noise, read noise, row noise, quantization noise, and out-of-model noise, respectively. 
Apart from the out-of-model noise, other noise components are sampled from specific distributions:
\begin{equation}
\begin{split}\label{eq:noise_Eq}
&N_{shot}+I \sim \mathcal{P}(\frac{I}{K})K,\\
&N_{read} \sim \TL(\lambda;\mu_c,\sigma_{\TL}),\\
&N_{row} \sim \mathcal{N}(0,\sigma_r),\\
&N_{quant} \sim U(-\frac{1}{2}, \frac{1}{2}),
\end{split}
\end{equation}
where $K$ denotes the overall system gain. 
Here, $\mathcal{P}$, $\mathcal{N}$, and $U$ represent Poisson, Gaussian, and uniform distributions, respectively. 
$\TL(\lambda; \mu, \sigma)$ stands for the Tukey-lambda distribution~\cite{joiner1971some} with shape $\lambda$, mean $\mu$, and standard deviation $\sigma$. 
Based on the assumption in ELD~\cite{wei2021physics}, a linear relationship governs the joint distribution of $(K, \sigma_{\TL})$ and $(K, \sigma_r)$, expressed as:
\begin{align}
\begin{split}\label{eq:2pair}
&\log(K)\sim U(\log(\hat{K}_{min}), \log(\hat{K}_{max})),\\
&\log(\sigma_{\TL})|\log(K)\sim \mathcal{N}(a_{\TL}\log(K)+b_{\TL},\hat{\sigma}_{\TL}),\\
&\log(\sigma_{r})|\log(K)\sim \mathcal{N}(a_{r}\log(K)+b_{r},\hat{\sigma}_{r}),
\end{split}
\end{align}
where $\hat{K}_{min}$, $\hat{K}_{max}$ denotes the range of the overall system gain, determined by the minimal and maximum ISO value.
$a$, $b$, and $\hat{\sigma}$ indicate the line's slope, bias, and an unbiased estimator of the standard deviation, respectively.
In this context, a camera can be approximated as a ten-dimensional coordinate $\mathcal{C}$:
\begin{align}\label{eq:coor}
\mathcal{C} = (\hat{K}_{min}, \hat{K}_{max}, \lambda, \mu_c, a_{\TL}, b_{\TL}, \hat{\sigma}_{\TL}, a_{r}, b_{r}, \hat{\sigma}_{r}).
\end{align}

Existing methods predominantly rely on explicit calibration to determine the coordinate $\mathcal{C}$, especially the linear relationship.
It is a process characterized by intensive labor and a substantial domain gap (\ie the gap between simulated noise and real noise). 
Moreover, the entanglement between neural networks and cameras requires repeated explicit calibration and training.
In our implementation, these distributions and linear relationships are defined similarly to ELD~\cite{wei2021physics}. 
However, we can also employ more advanced noise models as replacements to achieve theoretically superior performance.

We aim to streamline the complex calibration process and mitigate the strong coupling between networks and cameras. 
Additionally, we address the out-of-model noise comprehensively, a task facilitated by the structural modifications introduced in the RepNR block. 
Our motivation is to compel the network to function as a swift adapter~\cite{ravi2016optimization,finn2017model}.

\begin{algorithm}[b] \label{alg1} 
  \caption{Pre-training pipeline in \framework} 
  \begin{algorithmic}
    \REQUIRE model $\Phi, m,\mathcal{S},$ clean dataset $ D$
    \STATE $\Phi_{\text{pre}} \gets$ insert-multi-CSA($\Phi$) 
    \STATE $\{c_k\}_{k=1}^m \gets$ generate-virtual-camera($\mathcal{S}$)
    \WHILE{not converged}
      \STATE Sample mini-batch $ x_i \sim D$
      \STATE $k \gets $ random$(1,m)$
      \STATE $\Tilde{x_i} \gets$ augment$(c_k, x_i)$
      \STATE $\Phi_{\text{pre},k} \gets$ select-CSA($\Phi_{\text{pre}}, k$)
      \STATE train$(\Phi_{\text{pre},k}, \{\Tilde{x_i}, x_i\})$
    \ENDWHILE
  \end{algorithmic} 
\end{algorithm}

\subsection{Pre-train with Camera-Specific Alignment}
\label{sec:pretrain}
\smallsec{Preprocessing.}
We initiate the pre-training stage using virtual cameras to induce the network to function as a fast adapter. 
Given the number of virtual cameras $m$ and the parameter space (formulated as $\mathcal{S}$), for the $k$-th camera, we select the $k$-th $m$ bisection points for each parameter range and combine them to construct a virtual camera. 
Augmenting the data with synthetic noise, we can pre-train our network based on multiple virtual cameras, compelling the network to acquire common knowledge.

\smallsec{Camera-Specific Alignment.}
\label{CSA}
As depicted in \figref{fig:archs}, within the pre-training process, we introduce our Camera-Specific Alignment (CSA) module, which focuses on adjusting the distribution of input features. 
In the baseline model, a $3\times 3$ convolution followed by leaky-ReLU~\cite{xu2015empirical} constitutes the primary component. 
A multi-path alignment layer is inserted before each convolution of the network to align features from different virtual cameras into a shared space.
Each path represents the CSA corresponding to the $k$-th camera, aligning the $k$-th camera-specific feature distribution into a shared space. 
Let the feature of the $k$-th virtual camera be $F_k\in \mathcal{R}^{B\times C\times H\times W}$. 
Formally, the $k$-th branch contains a weight $W_k\in \mathcal{R}^{C}$ and a bias $b_k\in \mathcal{R}^{C}$, performing {\textbf{channel-wise}} linear projection, denoted by $Y=W_kF + b_k$. 
${W_k}$ are initialized as $\mathbf{1}$, and ${b_k}$ are initialized as $\mathbf{0}$, with no effect on the $3\times 3$ convolution at the beginning.

During training, data augmented by the noise of the $k$-th virtual camera is fed into the $k$-th path for alignment and a shared $3\times 3$ convolution for further processing. 
The detailed pre-training pipeline is described in \algref{alg1}.

\subsection{Fine-tune with Few-shot RAW Image Pairs}
\label{ft_few}
Following the pre-training process, the model is intended for deployment in realistic denoising tasks. 
We advocate for a few-shot strategy, specifically employing only 6 pairs (2 pairs for each of the three ratios) of raw images to fine-tune the pre-trained model. 
We assume that $3\times3$ convolutions have acquired sufficient capability to handle features aligned by CSAs. 
The convolutions remain frozen during subsequent fine-tuning to maximize the utilization of the model parameters obtained from pre-training.
For addressing real noise, we substitute the multi-branch CSA with a new CSA layer, denoted as CSA$^T$ (CSA for the target camera). 
Unlike the multi-branch CSA during pre-training, the CSA$^T$ layer is initialized by averaging the pre-trained CSAs for improved generalization. 
The CSA$^T$ followed by a $3\times 3$ convolution branch mentioned above is called the in-model noise removal branch (IMNR).

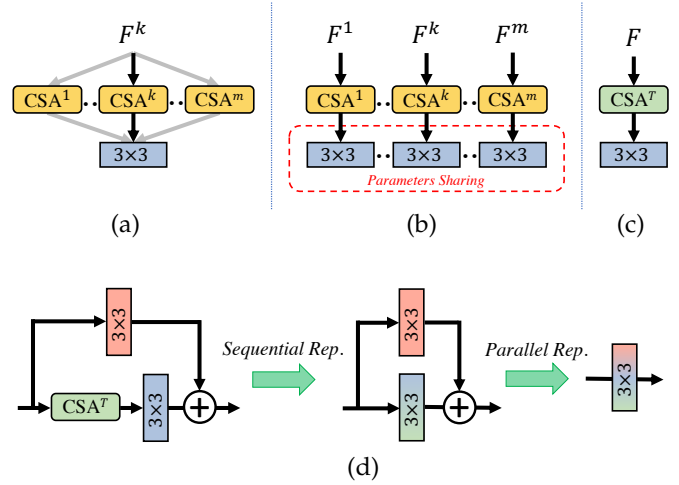
\begin{figure}[t!]
  \centering
  \begin{overpic}[width=0.48\textwidth]{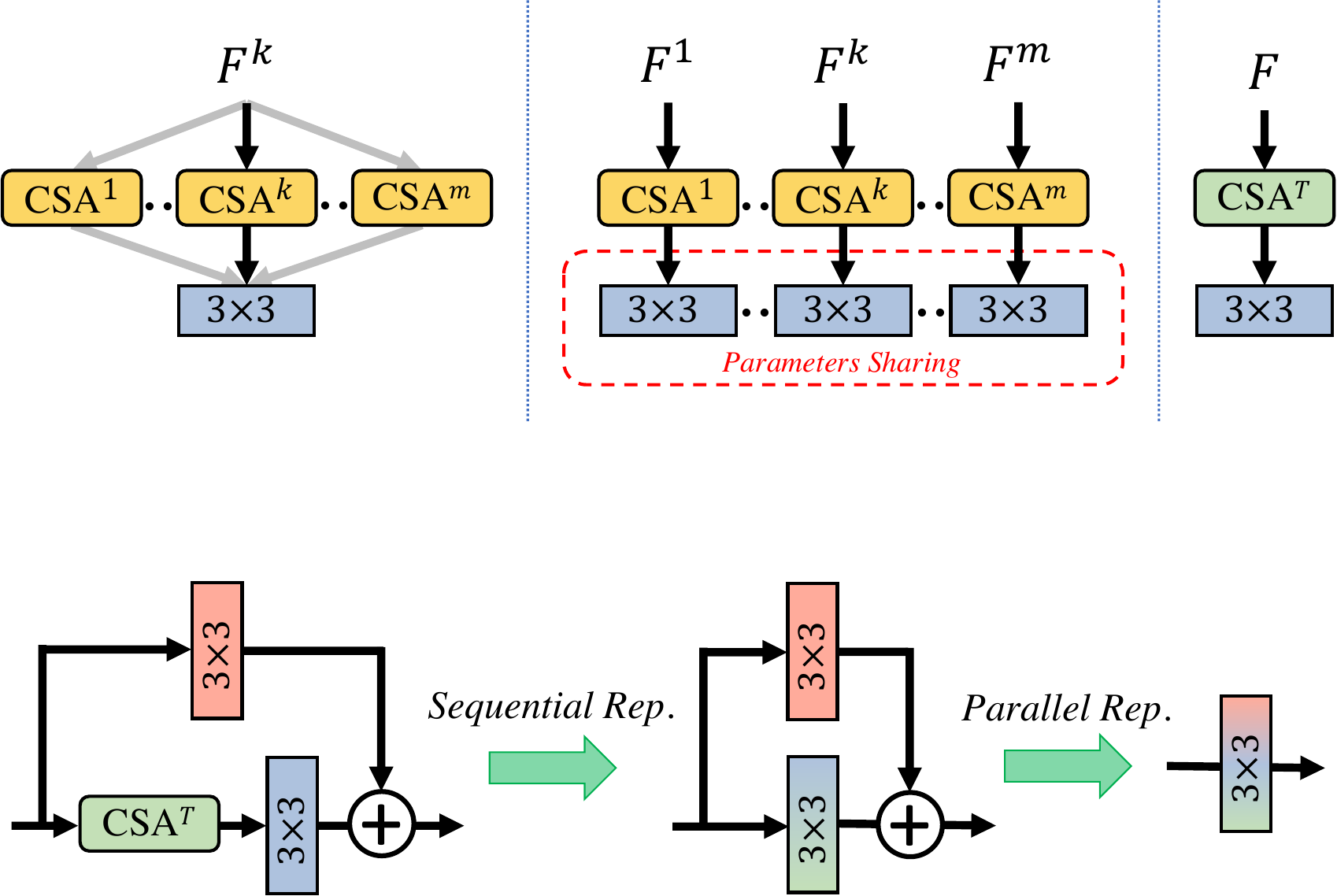}
    \put(15,32){(a)}
    \put(60,32){(b)}
    \put(92,32){(c)}
    \put(51, -5){(d)}
  \end{overpic}
  \vspace{3mm}
  \caption{
    Illustration for the initializing strategy of CSA$^T$ and the reparameterization process.
    (a) RepNR block during pre-training.
    (b) Our RepNR block can be seen as $m$ parameters sharing blocks, each for a specific virtual camera.
    (c) We initialize the CSA$^T$ by averaging the pre-trained CSAs, which can be considered model ensembling.
    (d) The reparameterization process during deployment. {\it{Rep.}} denotes reparameterize. We detailed the sequential reparameterization process in \secref{sec:deploy}.
  }
  \label{fig:rep+ensemble}
\end{figure}

Nevertheless, real noise encompasses the modeled part and some out-of-model noise. 
Since our CSA layer is specifically designed for aligning features augmented by synthetic noise, a gap still exists between real noise and the one that IMNR can handle (\ie $\epsilon$ in \eqnref{eq:nm}).
Therefore, we propose introducing an out-of-model noise removal branch (OMNR), to learn the gap between real noise and the modeled components. 
We treat the OMNR component as a parallel branch alongside the IMNR branch, due to previous research that has demonstrated the efficacy of parallel convolution branches in transfer and continual learning~\cite{zhang2022representation}. 
OMNR comprises only a $3\times3$ convolution, aiming to capture the structural characteristics of real noise from few-shot raw image pairs. 
Given the absence of prior information on the noise remainder $\epsilon$, we initialize the weights and bias of OMNR as a tensor of $\mathbf{0}$. 
Combining IMNR with OMNR yields the proposed RepNR block.
It is worth noting that it is more reasonable to first learn in-model noise and subsequently address out-of-model noise. 
Therefore, we divide the optimization process into two steps: initially training IMNR and subsequently training OMNR. 
Following this approach, iterations of two-step fine-tuning only account for 0.5$\%$ of the pre-training, rendering it highly feasible for practical implementation. 
The detailed fine-tuning pipeline is described in \algref{alg2}.

\smallsec{Analysis on the Initialization of CSA$^T$.}
As mentioned in \secref{ft_few}, we initialize CSA$^T$ by averaging the pre-trained CSAs in the multi-branch CSA layer. 
Given that every path shares the convolution in the multi-branch CSA, this initialization can be conceptualized as the ensemble of $m$ models, where $m$ is the number of paths, like (a)-(c) in \figref{fig:rep+ensemble}. 
According to studies~\cite{cha2021swad,izmailov2018averaging,xiao2023endpoints}, the weighted average of different models can significantly enhance the model's generalization. 
This aligns with our objective of generalizing the model to the target noisy domain.

Another rationale for this approach is that CSAs are largely determined by the coordinates $\mathcal{C}$. 
From this perspective, the average of different CSAs can be considered the center of gravity of these coordinates. 
Moreover, the coordinates of test cameras, both in SID~\cite{chen2018learning} and ELD~\cite{wei2021physics}, are encompassed in the parameter space $\mathcal{S}$. 
In such circumstances, averaging the pre-trained CSAs is a sound starting point. 
However, even if coordinates $\mathcal{C}$ are not in the pre-defined parameter space $\mathcal{S}$ (in our \dataset~dataset), \framework~could also achieve SOAT performance with a few more iterations during fine-tuning.

\begin{algorithm}[t]
  \caption{Fine-tuning and deploy pipeline in \framework}
  \label{alg2}
  \begin{algorithmic}
    \REQUIRE pre-trained model $\Phi_{\text{pre}}$, real dataset $D_{\text{real}}$
    \STATE $\Phi_{\text{ft}} \gets$ freeze-3$\times$3($\Phi_{\text{pre}}$) 
    \STATE $\Phi_{\text{ft}} \gets$ average-CSA($\Phi_{\text{ft}}$)
    \WHILE{not converged}
    \STATE Sample mini-batch pairs $\{x_i, y_i\} \sim D_{\text{real}}$
    \STATE train$(\Phi_{\text{ft}}, \{x_i, y_i\})$
    \ENDWHILE
    \STATE $\Phi_{\text{ft}} \gets$ freeze-IMNR($\Phi_{\text{ft}}$)
    \STATE $\Phi_{\text{ft}} \gets$ add-OMNR($\Phi_{\text{ft}}$)
    \WHILE{not converged}
    \STATE Sample mini-batch pairs $\{x_i, y_i\} \sim D_{\text{real}}$
    \STATE train$(\Phi_{\text{ft}}, \{x_i, y_i\})$
    \ENDWHILE
    \STATE $\Phi_{\text{final}}\gets$ deploy($\Phi_{\text{ft}}$) 
  \end{algorithmic}
\end{algorithm}

\subsection{Deploy}
\label{sec:deploy}
Upon completion of fine-tuning, the deployment of the model holds paramount importance for future applications. 
Directly substituting the $3\times 3$ convolution with our RepNR Block would inevitably increase the number of parameters and computational workload. 
However, it is noteworthy that our RepNR block solely comprises serial vs. parallel linear mappings. 
Additionally, the receptive field of each branch in the RepNR block is $3$. 
Therefore, employing the structural reparameterization technique~\cite{ding2019acnet,ding2021diverse,ding2021repvgg}, our RepNR block can be transformed into a plain $3\times 3$ convolution during deployment, as illustrated in \figref{fig:rep+ensemble} (d). 
This implies that our model incurs no additional costs in the application process and facilitates a fair comparison with other methods.
Regarding parallel reparameterization techniques, please refer to previous works~\cite{ding2019acnet,ding2021diverse,ding2021repvgg,ding2022repmlpnet,ding2023unireplknet}. Here, we primarily introduce the serial reparameterization techniques we employed.

\smallsec{Sequential Reparameterization.} The reparameterization process can be denoted as the following equation:
\begin{align}
\begin{split}
W_{\mathbf{rep}}&=\mathbf{diag}(W)\otimes W_{3\times 3},\\
b_{\mathbf{rep}}&=W_{3\times 3}\otimes \mathbf{pad}(b) + b_{3\times 3},
\end{split}
\end{align}
where $\mathbf{diag}$, $\mathbf{pad}$ denotes transform a $C$ dimensional vector into a $C\times C$ diagonal matrix and replicate-padding a $1\times1\times C$ dimensional vector into a $3\times 3\times C$ matrix respectively.
And $W$, $W_{3\times 3}$, and $W_{\mathbf{rep}}$ stand for the weight of the CSA, the $3\times 3$ convolution, and the reparameterized weight, respectively. And the $b_\ast$ are standing for the bias of the corresponding type. 

Since our CSA operator solely comprises $1\times1$ channel-wise operations, it is necessary to initially transform it into a regular $1\times1$ convolution using the $\mathbf{diag}$  operator during reparameterization. It is worth noting that such reparameterization can only approximate the $W_{\mathbf{rep}}$ and $b_{\mathbf{rep}}$. To ensure consistency during training and testing, we employed the online reparameterization technique~\cite{hu2022online}. It allows for reparameterization during training, which intends to save more GPU memories. However, our primary goal is to ensure consistency between training and testing utilizing the online reparameterization technique.
More details can be found in our Github repo~\cite{jin2023led_github}.

\section{Dark RAW Images (\dataset) Dataset}
\label{sec:dataset}

In this section, we will introduce the \dataset~dataset, details related to data collection (to guide the deployment of \framework~to any other cameras), and the availability and limitations of the data. Notice that, description in this section has been simplified as much as possible to facilitate a more comfortable and rapid deployment of \framework~on any other camera models.

\subsection{Overview of the \dataset~Dataset}

To further validate the effectiveness of \framework~across different cameras, we introduce the \dataset~dataset. 
Compared to existing datasets, our \dataset~dataset has the following advantages:
\begin{itemize}
  \setlength{\itemsep}{5pt} 
  \setlength{\parsep}{2pt}
  \item {\textbf{Multi-Camera Data}}: 
  To further demonstrate the effectiveness of \framework~across different cameras (corresponding to different noise parameters, coordinates $\mathcal{C}$), our dataset includes five distinct models not covered in existing datasets. 
  Additionally, \dataset~includes full-frame and APS-C format cameras with smaller sensor areas, often exhibiting stronger noise characteristics.
  \item {\textbf{Varied Illumination Settings}}: 
  The dataset contains data under five different illumination ratios ($\times 1$, $\times 10$, $\times 100$, $\times 200$, and $\times 300$), each representing varying levels of denoising difficulty.
  \item {\textbf{Dual ISO Configurations}}: 
  There are two different ISO settings for each scene and illumination setting. 
  These can be used not only for the fine-tuning stage of the \framework~method but also for testing the algorithm's robustness under different illumination settings.
\end{itemize}
In addition to the three highlighted points, the \dataset~dataset spans 30 indoor scenes, featuring diverse backgrounds and varying types and quantities of objects being photographed. 
It includes seven different ISO settings ranging from 200 to 6400. 
The hardest example in our dataset resembles the image captured at a ``pseudo" ISO up to 960,000 ($3200 \times 300$).
We captured a 5-image burst per setting to collect a broader range of noise samples for each ISO configuration under every illumination setting. 
This approach provides more test data pairs and lays the groundwork for burst raw image denoising in extremely dark environments.
Also, we captured data for explicit calibration to reproduce existing calibration-based methods for fully evaluation.

Most existing datasets directly use low ISO and long exposure images as ground truth because the noise produced at low ISO settings is often negligible in full-frame cameras. 
However, since our shooting equipment includes APS-C format cameras with smaller sensor areas, we need to additionally perform multi-frame averaging denoising on low ISO and long exposure images (4 frames in our implementations). 
Therefore, we collected a total of $(5*5*2)*5*30=7,500$ noisy images and $4*5*30=600$ images for creating $150$ ground-truths, comprising $(5*5*2)*5*30=7,500$ pairs of data for both training and evaluation.

\begin{figure*}
  \centering
  \includegraphics[width=\textwidth]{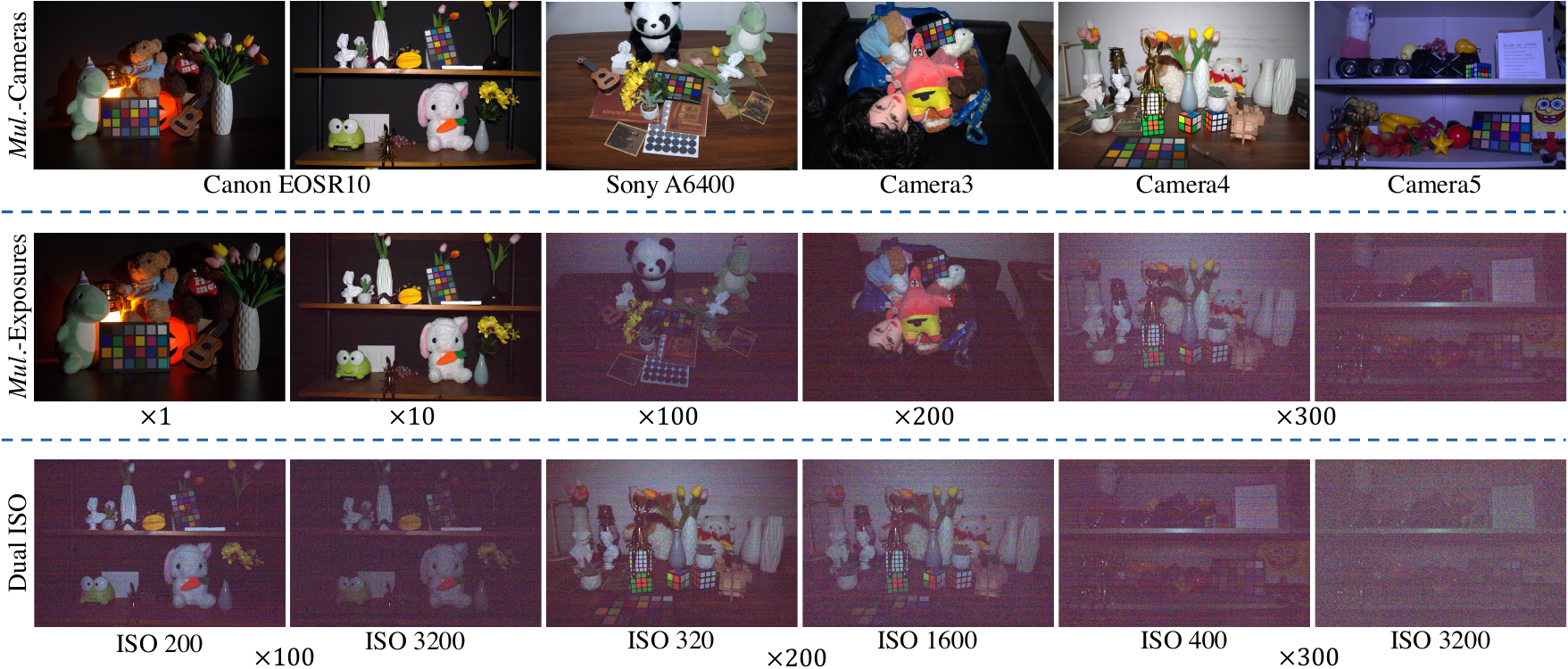}
  \vspace{-15pt}
  \caption{
    A thumbnail of our \dataset~dataset ({\textit{Zoom in for best view}}). It features 30 unique scenes, captured using 5 distinct camera models previously unrepresented in public datasets, under 5 varied lighting conditions (ranging from $\times 1$ to $\times 300$ ratios). 
    For each camera, scene, and lighting combination, we recorded images in dual ISO configurations to enhance the tuning of our \framework~(detailed in \secref{sec:discuss}), along with a burst of 5 images for expanded application. In total, \dataset~provides 7,500 paired images for both training and evaluative purposes. The visual results are amplified and post-processed with the ISP provided by RawPy~\cite{riechert2014rawpy}. Then, downsampled 4$\times$ to reduce file size.
  }\label{fig:multiraw_scenes}
\end{figure*}

\subsection{Instructions on Data Collection}

To ensure the quality of the dataset, special attention must be paid to lighting, alignment, and environmental factors during the shooting process:
\begin{itemize}
  \item {\textbf{Lighting}}: 
  To ensure consistent lighting conditions for the images, it is often necessary to supplement environmental lighting or adjust the aperture. 
  This allows for correct exposure in low ISO and long exposure scenarios.
  \item {\textbf{Alignment}}: 
  Remote control is essential to prevent misalignment issues. 
  Additionally, to avoid camera shake caused by the mechanical shutter during photography, the camera should be set to electronic shutter mode for shooting.
  \item {\textbf{Temperature}}: 
  To prevent the increase in camera temperature caused by continuous shooting (which typically leads to increased noise variance), it is necessary to set the interval between continuous shots to 5 seconds or more.
\end{itemize}
Moreover, to provide more information on signal-dependent noise (shot noise) for the fine-tuning of \framework, the scenes photographed should have a wide variety of colors.

\begin{table*}[tb]\small
  \tablestyle{1pt}{1.1}
  \caption{Quantitative results on the SID~\cite{chen2018learning} Sony subset. 
    The best result is in \best{bold}, whereas the second best one is in \second{underlined}. 
    The extra data requirements and iterations (K) are calculated when transferred to a new target camera.
    The DNN model-based methods require training noise generators for the target camera, resulting in larger iteration requirements.
    AINDNet* indicates that the AINDNet is pre-trained with our proposed noise model instead of AWGN. 
    It is worth noting that all methods except AINDNet are trained with the same UNet architecture, while we keep the AINDNet the same as their paper with almost twice the number of parameters compared to the UNet.
  }
  \vspace{-2mm}
  \label{tab:sid_comp}
\begin{tabular}{cccccccccc} \toprule
  \twRow{Categories} & \twRow{Methods} & \twRow{Extra Data Requirements} & \twRow{Iterations (K)} 
  & \twCol{$\times 100$} & \twCol{$\times 250$} & \twCol{$\times 300$} \\ \cmidrule{5-10} 
  & & & & PSNR & SSIM & PSNR & SSIM & PSNR & SSIM \\ \midrule
  \twRow{DNN Model Based} & Kristina~\etCite{monakhova2022dancing} & $\sim$1800 noisy-clean pairs 
  & 327.6 & 38.7799 & 0.9120 & 34.4924 & 0.7900 & 31.2971 & 0.6990 \\
  & NoiseFlow~\cite{abdelhamed2019noise} & $\sim$1800 noisy-clean pairs
  & 777.6 & 37.0200 & 0.8820 & 32.9457 & 0.7699 & 29.8068 & 0.6700 \\ \midrule
  \multirow{3}{*}{Calibration-Based} & Calibrated P-G & $\sim$300 calibration data 
  & 257.6 & 39.1576 & 0.8963 & 33.8929 & 0.7630 & 31.0035 & 0.6522 \\
  & ELD~\cite{wei2021physics} & $\sim$300 calibration data & 257.6 
  & \second{41.8271}& \second{0.9538}& 38.8492 & 0.9278 & 35.9402 & 0.8982 \\
  & Zhang~\etCite{zhang2021rethinking} & $\sim$150/$\sim$150 for calib./database  
  & 257.6 & 40.9232 & 0.9488 & 38.4397 & 0.9255 & 35.5439 & 0.8975 \\ \midrule
  \multirow{5}{*}{Real Data Based} & SID~\cite{chen2018learning} & $\sim$1800 noisy-clean pairs 
  & 257.6 & 41.7273 & 0.9531 & \second{39.1353}& \second{0.9304}& \best{37.3627} & \best{0.9341}  \\
  & Noise2Noise~\cite{lehtinen2018noise2noise} & $\sim$12000 noisy pairs 
  & 257.6 & 39.2769 & 0.8993 & 34.1660 & 0.7824 & 31.0991 & 0.7080 \\
  & AINDNet~\cite{kim2020transfer} & $\sim$300 noisy-clean pairs & 
  \abbest{1.5} & 40.5636 & 0.9194 & 36.2538 & 0.8509 & 32.2291 & 0.7397 \\
  & AINDNet* & $\sim$300 noisy-clean pairs 
  & \abbest{1.5} & 39.8052 & 0.9350 & 37.2210 & 0.9101 & 34.5615 & 0.8856 \\
  & \textit{\framework~(Ours)} & \abbest{6 noisy-clean pairs} 
  & \abbest{1.5} & \best{41.9842} & \best{0.9539} & \best{39.3419} & \best{0.9317} & \second{36.6728}& \second{0.9147}\\ \bottomrule
\end{tabular}
\vspace{-1mm}
\end{table*}

\newcommand{\tQuantHead}[1]{\toprule \twRow{Cam.} & \twRow{Ratio}& Calibrated P-G & ELD~\cite{wei2021physics} & {\it{\framework~(Ours)}} \\ & & PSNR/SSIM & PSNR/SSIM & PSNR/SSIM \\ \midrule \multirow{4}{*}{\rotatebox{90}{\makecell{#1}}}}
\newcommand{\tms}[1]{& $\times {#1}$}

\begin{table*}[tb]
  \tablestyle{5pt}{1.1}
  \caption{Quantitative results on four camera models, SonyA7S2, NikonD850, Canon EOS70D and Canon EOS700D, of the ELD~\cite{wei2021physics} dataset.
    The best result is denoted as \best{bold}. 
    The reasons for the significant performance improvement observed with Canon cameras are discussed in detail in~\secref{sec:discuss}. 
    All the metrics in this table are calculated with the last eight scenes in the ELD~\cite{wei2021physics} dataset, details in \label{sec:datasets}.
  }\vspace{-1mm}
  \label{tab:eld_comp}
\begin{tabular}{c|lccc} \tQuantHead{Sony \\ A7S2}
  \tms{1} & \best{54.3710}/\best{0.9977} & {52.8120}/0.9957 & 51.9547/{0.9968} \\
  \tms{10}  & 49.9973/0.9891 & {50.0152}/{0.9913} & \best{50.1762}/\best{0.9945} \\
  \tms{100} & 41.5246/0.8668 & {44.9865}/{0.9707} & \best{45.3574}/\best{0.9779} \\
  \tms{200} & 37.6866/0.7818 & {42.5440}/{0.9430} & \best{42.9747}/\best{0.9577} \\
  \bottomrule
\end{tabular} \hfill
\begin{tabular}{c|lccc} \tQuantHead{Nikon \\ D850}
  \tms{1} & {50.6207}/\best{0.9949} & 50.5628/0.9925 & \best{50.6222}/{0.9939} \\
  \tms{10}  & {48.3461}/0.9884 & \best{48.3667}/{0.9890} & 48.0684/\best{0.9894} \\
  \tms{100} & 42.2231/0.9046 & \best{43.6907}/{0.9634} & {43.5620}/\best{0.9667} \\
  \tms{200} & 39.0084/0.8391 & {41.3311}/{0.9364} & \best{41.3984}/\best{0.9482} \\
  \bottomrule
\end{tabular}\\ \vspace{1mm}
\begin{tabular}{c|lccc} \tQuantHead{Canon \\ EOS70D} 
  \tms{1}& 42.7352/0.9915 & 42.4305/0.9900 & \best{48.5063}/\best{0.9924} \\
  \tms{10} & 41.0061/0.9841 & 40.6364/0.9833 & \best{45.4415}/\best{0.9842} \\
  \tms{100} & 36.7007/0.8700 & 37.7944/0.9255 & \best{39.5491}/\best{0.9360} \\
  \tms{200} & 33.3459/0.7942 & 35.1554/0.8703 & \best{36.2362}/\best{0.8948} \\
  \bottomrule
\end{tabular} \hfill
\begin{tabular}{c|lccc} \tQuantHead{Canon \\ EOS700D} 
  \tms{1} & 42.0156/0.9900 & 41.9264/0.9881 & \best{47.7006}/\best{0.9910} \\
  \tms{10} & 40.7658/0.9791 & 40.5297/0.9758 & \best{44.8541}/\best{0.9815} \\
  \tms{100} & 36.7589/0.8697 & 36.9642/0.8937 & \best{38.3147}/\best{0.9206} \\
  \tms{200} & 34.3376/0.8063 & 34.9231/0.8534 & \best{35.1962}/\best{0.8717}  \\
  \bottomrule
\end{tabular}
\vspace{-2mm}
\end{table*}

\subsection{Dataset Application and Availability}

Our dataset will be used in the Few-shot RAW Image Denoising track at the CVPR 2024 workshop: Mobile Intelligent Photography \& Imaging. 
Following popular benchmarks, we fully release a subset of the data (about 20 scenes of the Canon EOSR10 and Sony A6400 camera models), along with a batch of test data. 
To prevent overfitting, we only make the images public, with the corresponding ground truths accessible via an online leaderboard on Google CodaLab~\cite{codalab_competitions_JMLR}.
A thumbnail of our \dataset~dataset is illustrated in \figref{fig:multiraw_scenes}.

\section{Experiments and Analysis}
\label{sec:exps}

This section offers a comprehensive description of our implementation, details the evaluation metrics and datasets  used, presents comparative experiments with other methods, and includes ablation studies to demonstrate the efficacy of our approach.

\subsection{Implementation Details}
\label{sec:implement}

Similar to most denoising methods~\cite{zamir2020cycleisp,cheng2021nbnet}, we utilize the $L1$ loss function as the training objective.
We adopt the same UNet~\cite{ronneberger2015u} architecture as previous methods for a fair comparison, with the distinction that we replace the convolution blocks inside the UNet with our proposed RepNR block.
As mentioned in \secref{sec:deploy}, the RepNR block can be structurally reparameterized into a simple convolution block without incurring additional computational costs.
We employ the same data preprocessing and optimization strategy as ELD~\cite{wei2021physics} during pre-training.
The raw images with long exposure time in the SID~\cite{chen2018learning} train subset are utilized for noise synthesis.
Concerning data preprocessing, we pack the Bayer images into 4 channels, followed by cropping the long exposure data with a patch size of $512\times 512$, non-overlapping, step $256$, thereby increasing the iterations of one epoch from $161$ to $1288$.
Our implementation is based on PyTorch~\cite{paszke2017automatic} and MindSpore~\cite{mindsporeai2023mindspore}. We train the models for 200 epochs (257.6K iterations) using the Adam optimizer~\cite{kingma2014adam} with $\beta_1=0.9$ and $\beta_2=0.999$ for optimization, without applying weight decay.
The initial learning rate is set to $10^{-4}$ and is halved at the 100th epoch (128.8K iterations) before being further reduced to $10^{-5}$ at the 180th epoch (231.84K iterations).

During fine-tuning, we initially freeze the $3\times 3$ convolution and average the multi-branch CSA to initialize CSA$^T$.
We first train CSA$^T$ until convergence, which constitutes the implicit calibration process we propose. After CSA$^T$ has converged, we introduce the out-of-model noise removal branch (a parallel $3\times 3$ convolution) and freeze all the remaining parameters in our network, as depicted in \figref{fig:archs}~\ding{175}. Subsequently, we train the OMNR until convergence. Different datasets require varying iterations and learning rates, the details of which will be described in \secref{sec:datasets}.
After completing the training process, we deploy our model by reparameterizing the RepNR blocks into convolutions.

\subsection{Evaluation Metrics and Datasets}

PSNR and SSIM~\cite{wang2004image} are utilized as quantitative evaluation metrics for pixel-wise and structural assessment.
It's important to note that the pixel value of low-light raw images usually lies in a smaller range than sRGB images, typically $[0, 0.5]$ after normalization. This can result in a lower mean square error and higher PSNR.
We evaluated our proposed \framework~on 3 RAW-based denoising datasets, namely SID~\cite{chen2018learning}, ELD~\cite{wei2021physics} and our proposed \dataset.

\smallsec{SID~\cite{chen2018learning} dataset.}
The SID~\cite{chen2018learning} dataset exclusively comprises the Sony A7S2 camera model, yet its test scenes are highly diverse, effectively demonstrating the algorithm's efficacy to the greatest extent. Consequently, a substantial number of ablation experiments are based on this dataset.
We randomly selected two pairs of data for each additional digital gain ($\times 100$, $\times 250$, and $\times 300$), in a total of six pairs, as the few-shot training datasets.
Since the coordinate $\mathcal{C}$ (first mentioned in \eqnref{eq:coor}) of the Sony A7S2 is already included in our pre-defined parameter space $\mathcal{S}$, the required training strategy can be relatively mild.
We initially fine-tuned CSA$^T$ using a learning rate of $10^{-4}$ for 1K iterations. Subsequently, we fine-tune the OMNR branch for 500 iterations using a learning rate of $10^{-5}$.

\smallsec{ELD~\cite{wei2021physics} dataset.}
The ELD~\cite{wei2021physics} dataset encompasses four camera models: Sony A7S2, Nikon D850, Canon EOS70D, and Canon EOS700D. 
We used the paired raw images of the first two scenarios for fine-tuning the pre-trained network, while the remaining eight scenarios were used for evaluation.
All the metrics in \tabref{tab:eld_comp} are calculated across the eight scenes for fair comparison.
On the ELD~\cite{wei2021physics} dataset, since the four cameras' coordinate $\mathcal{C}$s are all included in our pre-defined parameter space $\mathcal{S}$, the training strategy is the same as for the SID~\cite{chen2018learning} dataset.

\smallsec{\dataset~dataset.}
The \dataset~dataset includes five camera models not previously mentioned: Sony A6400, Canon EOSR10, and three other cameras.
Given that this dataset is intended for few-shot raw image denoising, we directly use its training set for fine-tuning. 
The training strategy on the \dataset~dataset may be somewhat aggressive because the coordinate $\mathcal{C}$s of the 5 camera models in \dataset~dataset are not included in our pre-defined parameter space $\mathcal{S}$. However, This would fully verify the effectiveness of our proposed \framework~on unseen camera models. During the fine-tuning process, we adopted the SGDR~\cite{loshchilov2016sgdr} learning rate decay strategy. Initially, CSA$^T$ is trained with a learning rate from $2\times 10^{-4}$ to $5\times 10^{-5}$ for 1K iterations for rapid convergence. Subsequently, the OMNR is trained for 2K iterations with a learning rate from $10^{-4}$ to $10^{-5}$.

\begin{table*}[tb]\small
  \tablestyle{2pt}{1.1}
\caption{
  Quantitative results on the five different camera models, Canon EOSR10, Sony A6400, and three other camera models, of the proposed \dataset~dataset. The best result is in \best{bold}. Time denotes the training time on a single Nvidia Geforce 3090 GPU with training strategy declared in \secref{sec:implement}. For {\textit{\framework}} and AINDNet~\cite{kim2020transfer}, Time denotes the training time of the fine-tuning stage (only when deploying to new camera models.). AINDNet* indicates that the AINDNet is pre-trained with our proposed noise model instead of AWGN. All methods except AINDNet are trained with the same UNet architecture, while we keep the AINDNet the same as their paper with almost twice the number of parameters compared to the UNet.
  Please note that these metrics were calculated across all scenarios of the proposed \dataset~dataset.
}
\label{tab:multiraw_comp}
\begin{center}
\begin{tabular}{c||l|ccccccccccccccc}
\toprule
\twRow{Camera}       & \twRow{Ratio} 
    & \multicolumn{3}{c}{P-G}                     
    & \multicolumn{3}{c}{AINDNet*~\cite{kim2020transfer}} 
    & \multicolumn{3}{c}{ELD~\cite{wei2021physics}} 
    & \multicolumn{3}{c}{Zhang~\etal~\cite{zhang2021rethinking}} 
    & \multicolumn{3}{c}{\it{\framework~(Ours)}}     \\
    &                        
    & PSNR    & SSIM   & Time                     
    & PSNR    & SSIM   & Time                     
    & PSNR    & SSIM   & Time                     
    & PSNR    & SSIM   & Time                     
    & PSNR    & SSIM   & Time                     \\\midrule
\multirow{5}{*}{\shortstack{Canon\\EOSR10}}  
    & $\times 1$             
      & 45.5070 & 0.9895 & \multirow{5}{*}{\shortstack[l]{4h\\35m\\27s}} 
      & 42.8885 & 0.9749 & \multirow{5}{*}{\shortstack[l]{15m\\01s}} 
      & 45.4837 & 0.9786 & \multirow{5}{*}{\shortstack[l]{4h\\37m\\11s}} 
      & 45.4036 & 0.9865 & \multirow{5}{*}{\shortstack[l]{4h\\29m\\12s}}
      & \best{48.6290} & \best{0.9918} & {\textbf{\multirow{5}{*}{\shortstack[l]{7m\\17s}}}} \\
    & $\times 10$            
      & 44.7179 & \best{0.9847} &                          
      & 41.8977 & 0.9670 &                          
      & 43.4092 & 0.9601 &                          
      & 43.9946 & 0.9803 &                          
      & \best{46.3750} & 0.9842 &                          \\
    & $\times 100$           
      & 39.8212 & 0.9064 &                          
      & 39.2519 & 0.9391 &                          
      & 40.6755 & 0.9310 &                          
      & 41.2814 & \best{0.9594} &                          
      & \best{41.8574} & 0.9547 &                          \\
    & $\times 200$           
      & 37.0122 & 0.8130 &                          
      & 38.3639 & 0.9279 &                          
      & 40.3582 & 0.9439 &                          
      & 40.1521 & \best{0.9486} &                          
      & \best{40.8654} & 0.9456 &                          \\
    & $\times 300$           
      & 34.5953 & 0.7769 &                          
      & 35.7965 & 0.8700 &                          
      & 37.7036 & \best{0.8987} &                          
      & 37.6117 & 0.8967 &                          
      & \best{37.7800} & 0.8972 &                          \\\midrule
\multirow{5}{*}{\shortstack{Sony\\A6400}}    
    & $\times 1$             
      & \best{49.3146} & 0.9934 & \multirow{5}{*}{\shortstack[l]{4h\\23m\\15s}} 
      & 43.5193 & 0.9750 & \multirow{5}{*}{\shortstack[l]{15m\\15s}} 
      & 48.9889 & 0.9927 & \multirow{5}{*}{\shortstack[l]{4h\\39m\\27s}} 
      & 48.3114 & 0.9913 & \multirow{5}{*}{\shortstack[l]{4h\\29m\\32s}} 
      & 49.0211 & \best{0.9936} & {\textbf{\multirow{5}{*}{\shortstack[l]{7m\\19s}}}} \\
    & $\times 10$            
      & \best{47.7593} & \best{0.9880} &                          
      & 42.7484 & 0.9677 &                          
      & 47.1114 & 0.9835 &                          
      & 46.6079 & 0.9843 &                          
      & 47.4265 & \best{0.9880} &                          \\
    & $\times 100$           
      & 43.6363 & 0.9415 &                          
      & 41.0480 & 0.9531 &                          
      & 43.1836 & 0.9346 &                          
      & 43.3121 & 0.9505 &                          
      & \best{43.7688} & \best{0.9613} &                          \\
    & $\times 200$           
      & 41.3958 & 0.9131 &                          
      & 39.8725 & 0.9383 &                          
      & 42.0199 & 0.9204 &                          
      & 42.1055 & 0.9379 &                          
      & \best{42.5766} & \best{0.9562} &                          \\
    & $\times 300$           
      & 38.1028 & 0.8427 &                          
      & 38.0563 & 0.9098 &                          
      & 39.5744 & 0.8873 &                          
      & 40.2146 & 0.9169 &                          
      & \best{40.3370} & \best{0.9381} &                          \\\midrule
\multirow{5}{*}{Camera3}      
    & $\times 1$             
      & 41.1760 & 0.9798 & \multirow{5}{*}{\shortstack[l]{4h\\36m\\23s}} 
      & 40.7700 & 0.9594 & \multirow{5}{*}{\shortstack[l]{15m\\15s}} 
      & 40.5599 & 0.9796 & \multirow{5}{*}{\shortstack[l]{4h\\38m\\12s}} 
      & 42.0061 & 0.9790 & \multirow{5}{*}{\shortstack[l]{4h\\30m\\33s}} 
      & \best{42.3091} & \best{0.9816} & {\textbf{\multirow{5}{*}{\shortstack[l]{7m\\13s}}}} \\
    & $\times 10$            
      & 40.0307 & 0.9677 &                          
      & 39.4657 & 0.9420 &                          
      & 39.6185 & 0.9666 &                          
      & 40.4674 & 0.9672 &                          
      & \best{40.7769} & \best{0.9700} &                          \\
    & $\times 100$           
      & 36.2148 & 0.8938 &                          
      & 36.1391 & 0.8914 &                          
      & 36.7027 & 0.9138 &                          
      & 37.2370 & 0.9280 &                          
      & \best{37.4741} & \best{0.9311} &                          \\
    & $\times 200$           
      & 34.3638 & 0.8487 &                          
      & 35.1045 & 0.8783 &                          
      & 35.2796 & 0.8791 &                          
      & \best{36.0706} & 0.9045 &                          
      & 36.0443 & \best{0.9130} &                          \\
    & $\times 300$           
      & 30.4170 & 0.7663 &                          
      & 31.4775 & 0.7760 &                          
      & 31.8913 & 0.8211 &                          
      & 32.8985 & 0.8532 &                          
      & \best{33.0504} & \best{0.8561} &                          \\\midrule
\multirow{5}{*}{Camera4}      
    & $\times 1$             
      & 49.2394 & 0.9942 & \multirow{5}{*}{\shortstack[l]{4h\\36m\\20s}} 
      & 43.7557 & 0.9705 & \multirow{5}{*}{\shortstack[l]{15m\\08s}} 
      & 47.9876 & 0.9924 & \multirow{5}{*}{\shortstack[l]{4h\\38m\\15s}} 
      & 47.4546 & 0.9887 & \multirow{5}{*}{\shortstack[l]{4h\\30m\\30s}} 
      & \best{50.1183} & \best{0.9945} & {\textbf{\multirow{5}{*}{\shortstack{7m\\19s}}}} \\
    & $\times 10$            
      & 47.6744 & \best{0.9895} &                          
      & 42.9754 & 0.9636 &                          
      & 46.3897 & 0.9811 &                          
      & 45.8446 & 0.9768 &                          
      & \best{47.7583} & \best{0.9895} &                          \\
    & $\times 100$           
      & 41.9510 & 0.9335 &                          
      & 39.8534 & 0.9360 &                          
      & \best{42.4956} & 0.9537 &                          
      & 42.0030 & 0.9540 &                          
      & 41.9648 & \best{0.9587} &                          \\
    & $\times 200$          
      & 40.5930 & 0.9230 &                          
      & 38.7384 & 0.9294 &                          
      & \best{41.0072} & 0.9463 &                          
      & 40.3252 & 0.9354 &                          
      & 40.5241 & \best{0.9503} &                          \\
    & $\times 300$           
      & 36.6494 & 0.8391 &                          
      & 36.2330 & 0.8915 &                          
      & 38.5018 & 0.9108 &                          
      & \best{38.6361} & \best{0.9231} &                          
      & 38.1756 & 0.9209 &                          \\\midrule
\multirow{5}{*}{Camera5}      
    & $\times 1$             
      & \best{48.6019} & \best{0.9928} & \multirow{5}{*}{\shortstack[l]{4h\\24m\\03s}} 
      & 42.8059 & 0.9713 & \multirow{5}{*}{\shortstack[l]{14m\\58s}} 
      & 47.1503 & 0.9874 & \multirow{5}{*}{\shortstack[l]{4h\\18m\\44s}} 
      & 46.0550 & 0.9868 & \multirow{5}{*}{\shortstack[l]{4h\\29m\\52s}} 
      & 46.9796 & 0.9897 & {\textbf{\multirow{5}{*}{\shortstack[l]{7m\\16s}}}} \\
    & $\times 10$           
      & 43.4577 & 0.9134 &                          
      & 41.6037 & 0.9545 &                          
      & 43.5000 & 0.9627 &                          
      & 43.9310 & 0.9749 &                          
      & \best{44.5822} & \best{0.9753} &                          \\
    & $\times 100$           
      & 36.4346 & 0.7930 &                          
      & 38.1994 & 0.9081 &                          
      & 39.6707 & 0.9040 &                          
      & 39.9786 & 0.9321 &                          
      & \best{41.3606} & \best{0.9478} &                          \\
    & $\times 200$           
      & 32.6378 & 0.7228 &                          
      & 36.4481 & 0.8836 &                          
      & 37.3455 & 0.8712 &                          
      & 37.6322 & 0.9017 &                          
      & \best{39.8046} & \best{0.9307} &                          \\
    & $\times 300$           
      & 29.2045 & 0.6537 &                          
      & 32.9607 & 0.8229 &                          
      & 34.5113 & 0.8179 &                          
      & 33.9278 & 0.8524 &                          
      & \best{36.4322} & \best{0.8922} &                          \\
\bottomrule
\end{tabular}
\end{center}
\vspace{-4mm}
\end{table*}

\begin{figure*}
  \centering
  \vspace{5mm}
  \begin{overpic}[width=0.99\textwidth]{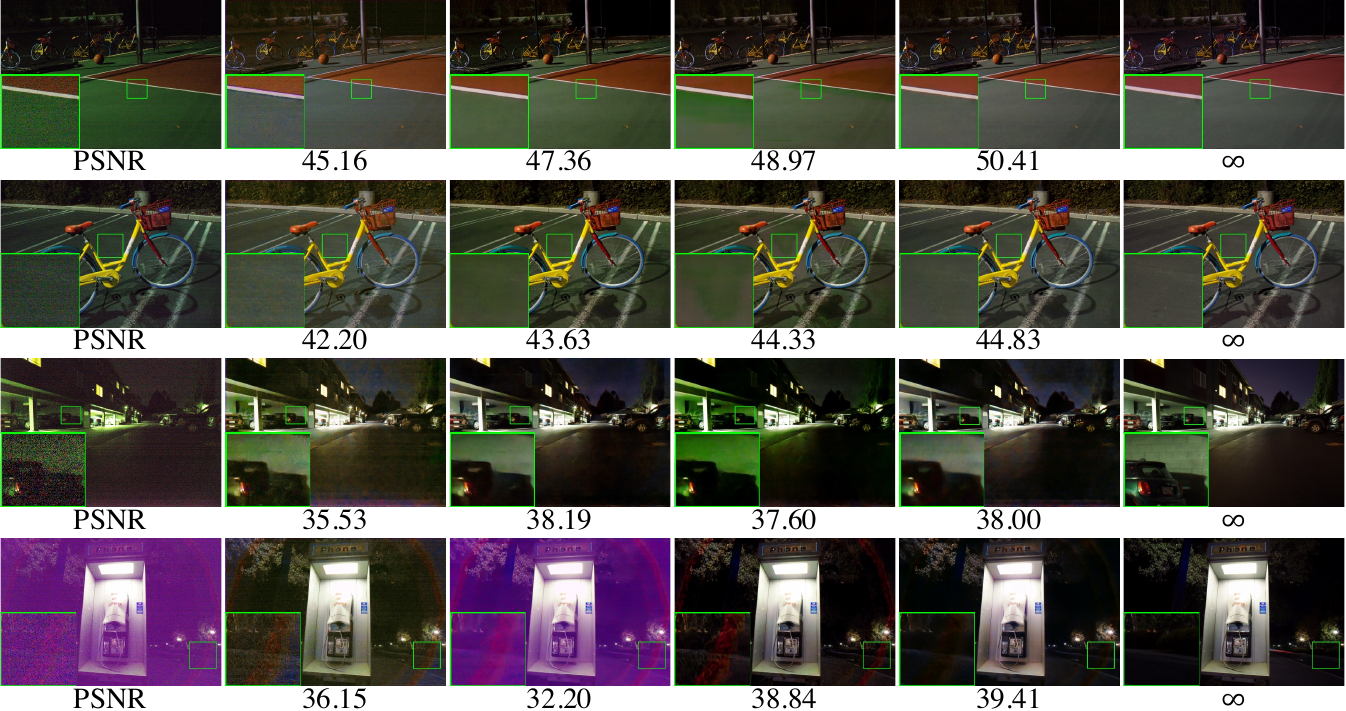}
      \put(5.8,53.5){{Input}}
      \put(18.9,53.5){{AINDNet~\cite{kim2020transfer}}}
      \put(34.6,53.5){{Zhang~\etCite{zhang2021rethinking}}}
      \put(55,53.5){{ELD~\cite{wei2021physics}}}
      \put(70.2,53.5){{{\bf{\framework~(Ours)}}}}
      \put(90.5,53.5){GT}
  \end{overpic}
  \caption{
    Visual comparison between our \framework~and other state-of-the-art methods on the SID~\cite{chen2018learning} dataset ({\it{Zoom-in for best view}}). We amplified and post-processed the input images with the same ISP as ELD~\cite{wei2021physics}.
  }
  \vspace{-3mm}
  \label{fig:sid_compare}
\end{figure*}

\subsection{Comparison with State-of-the-art Methods}

We assess the performance of our \framework~on three distinct datasets: the Sony subset of SID~\cite{chen2018learning}, the ELD dataset~\cite{wei2021physics}, and the 5 subsets in our \dataset~dataset. This evaluation aims to gauge the generalization capabilities of \framework~across outdoor and indoor scenes and across more camera models, respectively.
\framework~is benchmarked against state-of-the-art raw denoising methods designed for extremely low-light environments. These comparative analyses include:
\begin{itemize}
  \item{\bf{DNN model-based methods}}: Exemplars in this category encompass the approaches presented by Kristina \etCite{monakhova2022dancing} and NoiseFlow~\cite{abdelhamed2019noise}. These methodologies initially undergo training on paired real raw images, enabling them to learn the intricacies of noise generation specific to a particular camera. However, they may necessitate additional iterations when applied to a novel camera model.
  \item{\bf{Calibration-based methods}}: This classification encompasses ELD~\cite{wei2021physics}, the approach proposed by Zhang \etCite{zhang2021rethinking}, and Calibrated P-G. Noteworthy is the requirement for a time-intensive and laborious calibration process intrinsic to these methods.
  \item{\bf{Real data-based methods}}: Techniques falling under this category involve training with various data pairings, such as noisy-clean pairs (SID~\cite{chen2018learning}), noisy-noisy pairs (Noise2Noise~\cite{lehtinen2018noise2noise}), and transfer learning as demonstrated by AINDNet~\cite{kim2020transfer}.
\end{itemize}
The denoising network for all methods above is trained under identical settings, following the parameters outlined in ELD~\cite{wei2021physics}. This standardization ensures a fair and consistent basis for comparison, as elucidated in \secref{sec:implement}.

\smallsec{Quantitative Evaluation.}
As demonstrated in \tabref{tab:sid_comp}, \tabref{tab:eld_comp} and \tabref{tab:multiraw_comp}, our approach surpasses previous calibration-based methods in denoising performance under extremely low-light conditions.
The disparity between synthetic and real noise is exacerbated with a substantial ratio ($\times 250$ and $\times 300$), resulting in diminished performance during training with synthetic noise. 
This is exemplified in comparing ELD~\cite{wei2021physics} and SID~\cite{chen2018learning}.
Moreover, DNN model-based methods often exhibit more significant discrepancies than calibration-based methods, with Kristina \etCite{monakhova2022dancing} failing to account for different system gains. Our method mitigates this discrepancy by fine-tuning with few-shot real data, achieving superior performance under $\times 100$ and $\times 250$ digital gain, as detailed in \tabref{tab:sid_comp}.
AINDNet~\cite{kim2020transfer} also demonstrates enhanced performance under extremely dark scenes, benefitting from a noise model with reduced deviation. 
Notably, the noise model deviation has minimal impact on denoising efficacy under small additional digital gain, even may enhance performance, as illustrated in \tabref{tab:eld_comp}.
Discussions related to this phenomenon can be found in \secref{sec:discuss}.
Significantly, our method exhibits superiority under extremely low-light scenes, even across different camera models. 
Additionally, when compared to alternative methods, \framework~introduces lower training costs in terms of data requirements, training iterations, and training time.

\begin{figure}
  \centering
  \vspace{3.2mm}
  \begin{overpic}[width=0.48\textwidth]{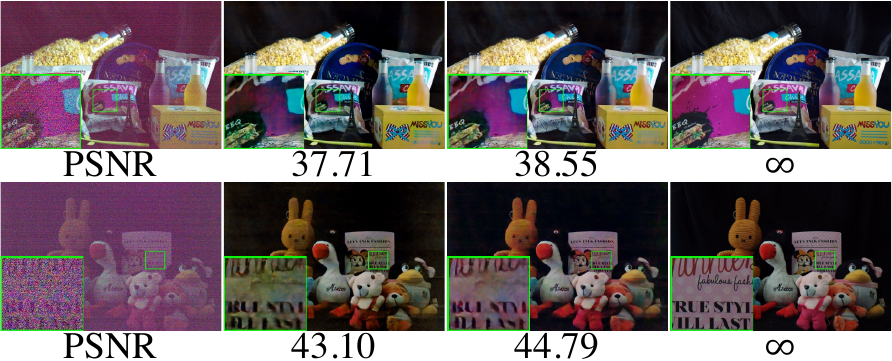}
    \put(8,41.5){Input}
    \put(30.8,41.5){ELD~\cite{wei2021physics}}
    \put(52.4,41.5){{\bf{\framework~(Ours)}}}
    \put(84.2,41.5){GT}
  \end{overpic}
  \vspace{-2mm}
  \caption{
    Visual comparison on the ELD~\cite{wei2021physics} dataset ({\it{Zoom-in for best view}}). 
  }
  \vspace{-2mm}
  \label{fig:eld_compare}
\end{figure}

\begin{figure*}
  \centering
  \vspace{5mm}
  \begin{overpic}[width=0.98\textwidth]{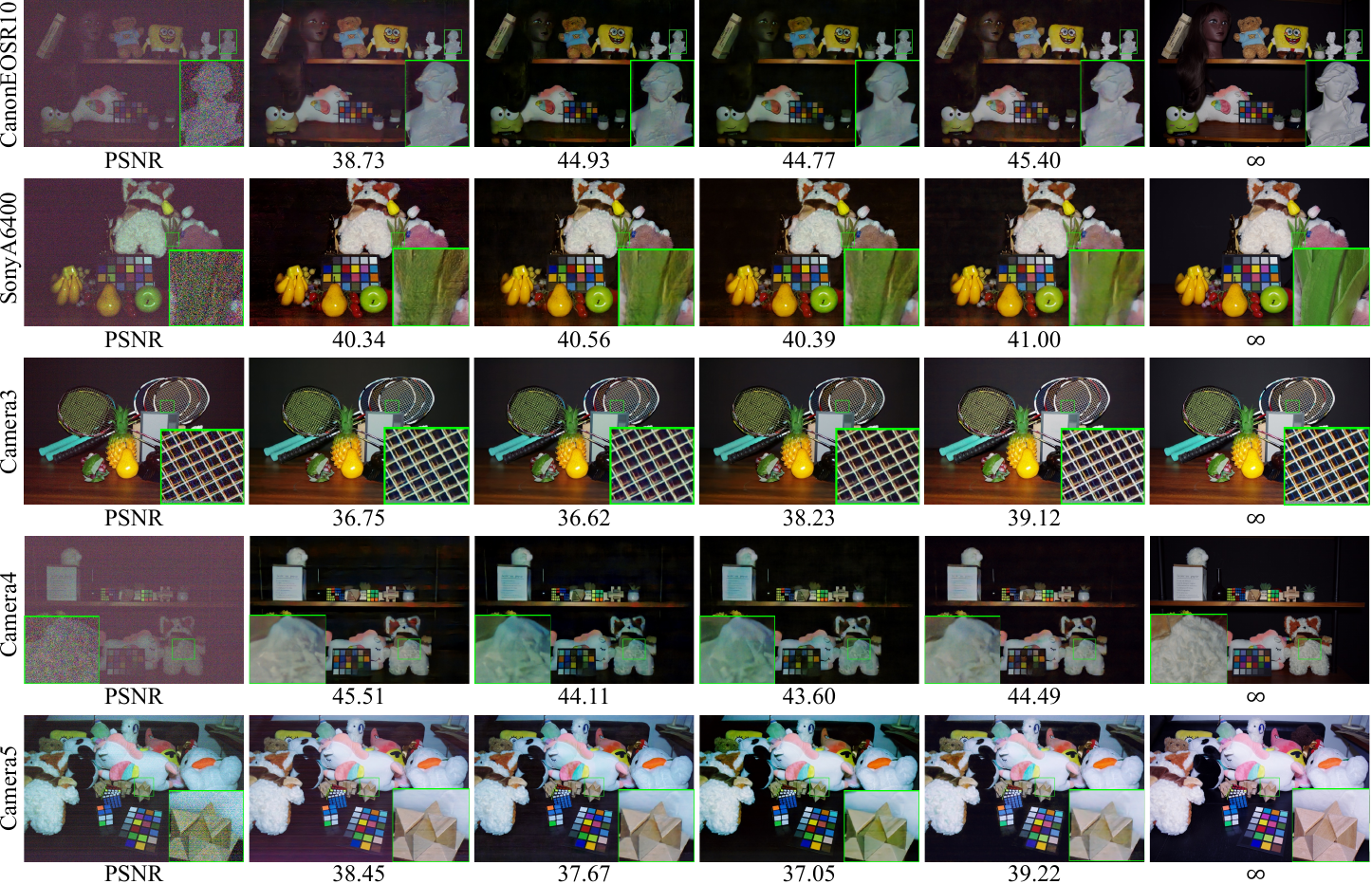}
      \put(7.5,65){{Input}}
      \put(24.5,65){{P-G}}
      \put(39,65){{ELD~\cite{wei2021physics}}}
      \put(52,65){{Zhang~\etCite{zhang2021rethinking}}}
      \put(70.5,65){{{\bf{\framework~(Ours)}}}}
      \put(90.5,65){GT}
  \end{overpic}
  \caption{
    Visual comparison between our \framework~and other state-of-the-art calibration-based methods on our proposed \dataset~dataset, along with 5 cameras ({\it{Zoom-in for best view}}). We amplified and post-processed the input images with the same ISP as ELD~\cite{wei2021physics}.
  }
  \label{fig:multiraw_compare}
\vspace{-7mm}
\end{figure*}
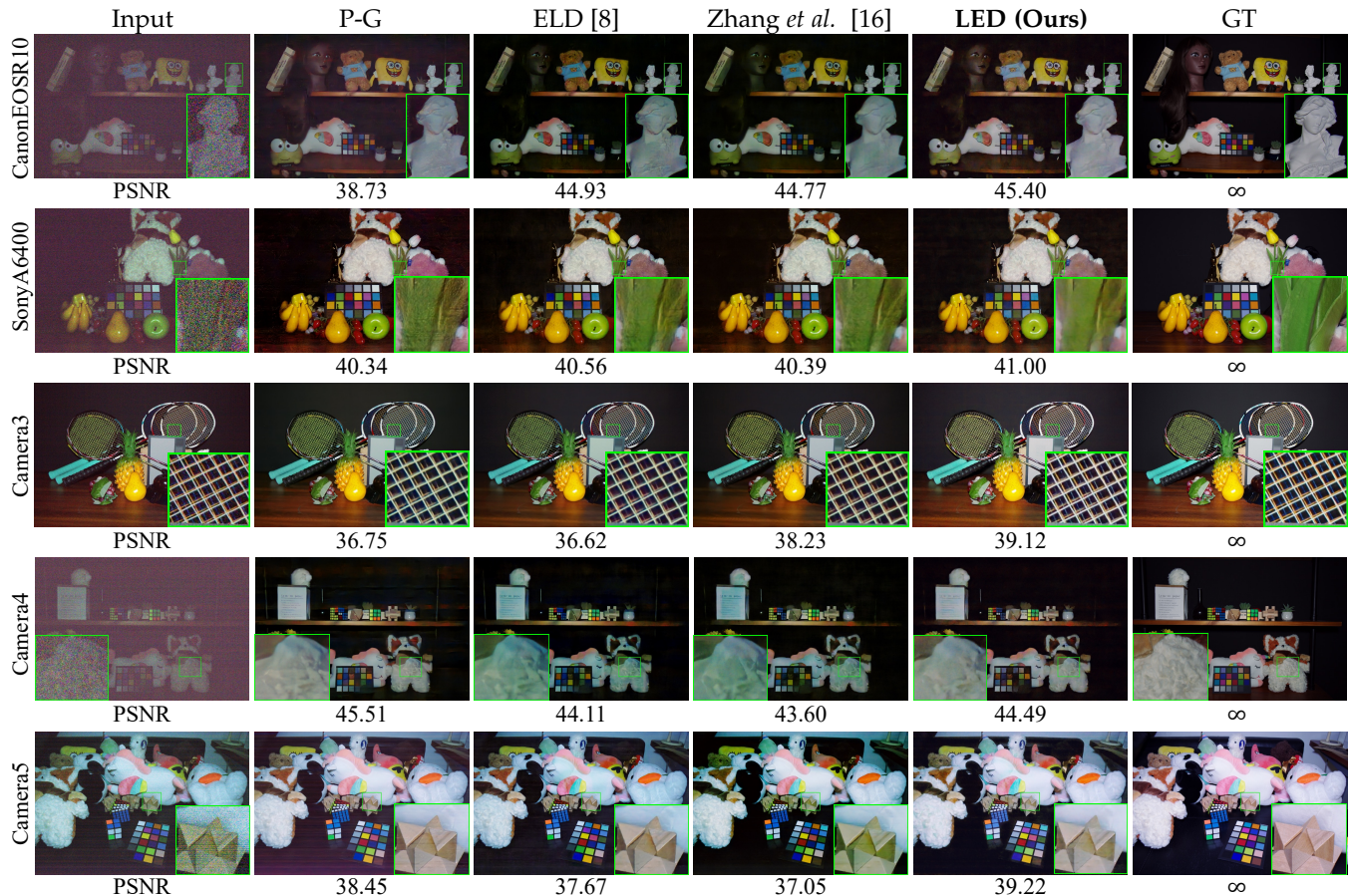

\smallsec{Qualitative Evaluation.}
The visual comparisons presented in \figref{fig:sid_compare}, \figref{fig:eld_compare} and \figref{fig:multiraw_compare} illustrate the performance of our method against other state-of-the-art approaches on the SID~\cite{chen2018learning}, ELD~\cite{wei2021physics} and \dataset ~datasets, respectively.
Under extremely low-light conditions, \framework~recovers more high-frequency information. 
As shown in Camera3 in \figref{fig:multiraw_compare}, \framework~is the only method to restore the strings of all three badminton rackets, especially the blue one.
Also, the presence of intense noise significantly disrupts the color tone. In \figref{fig:sid_compare}, input images exhibit noticeable green or purple color shifts, with many comparative methods struggling to restore the correct color tone. Leveraging implicit noise modeling and a diverse sampling space, \framework~efficiently reconstructs signals amidst severe noise interference, achieving accurate color rendering and preserving rich texture detail.
Moreover, other methods often fail to discern and address enlarged out-of-model noises, resulting in the corruption of the final image with fixed patterns or specific positional artifacts. In contrast, during the fine-tuning, \framework~learns to effectively eliminate these camera-specific noises, enhancing visual quality and demonstrating robustness against such challenges.

\setlength{\tabcolsep}{2pt}
\begin{table}[tb]\small
  \tablestyle{2pt}{1.1}
\begin{center}
\caption{
  Ablation studies on the RepNR block. 
  The provided metrics are with the fine-tuning, as shown in \ding{174} of \figref{fig:archs}.
}
\label{ab:rep}
% \vspace{-3mm}
\begin{tabular}{ccc||ccc}
\toprule
 \multicolumn{3}{c}{\textbf{Setting}} & $\times$100    & $\times$250    & $\times$300    \\
  U-net & CSA  & OMNR       & PSNR/SSIM    & PSNR/SSIM    & PSNR/SSIM    \\
\midrule
\checkmark & &     & 41.518/0.951 & 39.140/0.923 & 36.273/0.898 \\
 \checkmark &\checkmark &    & 41.866/\abbest{0.954} & 39.201/0.931 & 36.499/0.912 \\
\checkmark & \checkmark & \checkmark & \abbest{41.984}/\abbest{0.954} & \abbest{39.342}/\abbest{0.932} & \abbest{36.673}/\abbest{0.915}\\
\bottomrule
\end{tabular}
\end{center}
\vspace{-3mm}
\end{table}
\setlength{\tabcolsep}{8pt}

\subsection{Ablation Studies}
\vspace{-2mm}
\smallsec{Reparameterized Noise Removal Block.}
We conduct experiments to analyze the impact of different components in the Reparameterized Noise Removal Block (RepNR). As depicted in~\tabref{ab:rep}, our RepNR consistently demonstrates improved performance across three different ratios, with each component in the RepNR block contributing positively to the overall pipeline.

\smallsec{Pre-training with Advanced Strategy.}
As outlined in \tabref{tab:pretrain}, pre-training with the SGDR~\cite{loshchilov2016sgdr} optimizer and larger batch size (equivalent to the training strategy of PMN~\cite{feng2022learnability}) yields further performance improvements, all while maintaining the {\textbf{same fine-tuning}} (2 image pairs for each ratio and 1.5K iterations). This underscores the scalability of the proposed \framework.
Additionally, in comparison to LLD~\cite{cao2023physics}, \framework~demonstrates superior performance with minimal data and training costs.

\smallsec{Comparison between CSA and Other Normalization.}
A similar technique to our proposed one is to insert normalization layers in the network, which is relatively common in transfer learning scenarios.
To show the superiority of CSA compared with the usual method, we directly replace CSAs with different kinds of normalization layers to observe the difference. 
As shown in~\tabref{ab:diff_norm}, Alternatives are Instance-Normalization~\cite{ulyanov2016instance}, Layer-Normalization~\cite{ba2016layer}, and Batch-Normalization~\cite{ioffe2015batch} ($*$ denotes BN without running-mean and running-variance). 
Any normalization cannot achieve comparable performance to CSA. 
One main reason is that the value range of features is crucial to the denoising task. 
Normalization seriously destroys the value range of the feature and breaks its stability. 
On the contrary, CSA roughly maintains the original value range, preventing model performance from collapsing.

\begin{table}[tb]\small
  \tablestyle{5pt}{1.1}
\begin{center}
\caption{
  Ablation studies on the pre-training strategy. The notation with $\star$ indicates utilizing the same training strategy as PMN~\cite{feng2022learnability} for the denoiser. 
  At the same time, \framework$\star$ employs this strategy specifically for the pre-training stage and keeps the fine-tuning the same as before.
}
\label{tab:pretrain}
\begin{tabular}{c||ccc}
\toprule
\twRow{\textbf{Method}} & $\times$100    & $\times$250    & $\times$300    \\
& PSNR/SSIM    & PSNR/SSIM    & PSNR/SSIM    \\
\midrule
\framework & 41.984/0.954 & 39.342/0.932 & 36.673/0.915\\
ELD$\star$~\cite{wei2021physics} & 42.081/\abbest{0.955} & 39.461/0.934 & 36.870/0.920\\
LLD$\star$~\cite{cao2023physics} & 42.100/\abbest{0.955} & 39.760/0.933 & 36.760/0.912\\
\framework$\star$ & \abbest{42.396}/\abbest{0.955} & \abbest{39.843}/\abbest{0.939} & \abbest{36.997}/\abbest{0.923} \\
\bottomrule
\end{tabular}
\end{center}
\vspace{-1mm}
\end{table}
\setlength{\tabcolsep}{8pt}

\setlength{\tabcolsep}{6pt}
\begin{table}[tb]\small
\begin{center}
\caption{
  Ablation studies on the CSA. BN* denotes batch normalization with running mean and running variance.
}
\vspace{-1mm}
\label{ab:diff_norm}
\begin{tabular}{cccccc}
\toprule
Metric & CSA    & IN~\cite{ulyanov2016instance}      & LN~\cite{ba2016layer}      & BN~\cite{ioffe2015batch}      & BN*     \\
\midrule
PSNR        & \abbest{39.161} & 26.596 & 26.605 & 26.412 & 23.995 \\
SSIM        & \abbest{0.9322} & 0.5883 & 0.5938 & 0.6066 & 0.4186  \\
\bottomrule
\end{tabular}
\end{center}
\vspace{-3mm}
\end{table}
\setlength{\tabcolsep}{8pt}

\begin{figure}
  \centering
  \vspace{3mm}
  \begin{overpic}[width=0.48\textwidth]{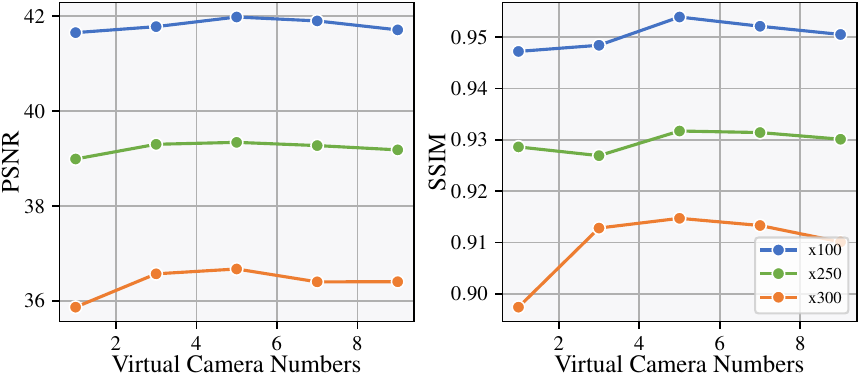}
  \end{overpic}
  \caption{
      Ablation studies on virtual camera numbers. 
      PSNR and SSIM reach the apex when the virtual camera number is 5. 
  }
  \label{fig:camera_number}
\end{figure}

\smallsec{Virtual Camera Number.}
We have done ablation studies on the virtual camera numbers of our proposed \framework. As shown in \figref{fig:camera_number},
\framework~achieves the best performance with five virtual cameras. 
Intuitive thought is that too few cameras will make it difficult for the model to learn common knowledge, while too many cameras significantly increase the difficulty of the model learning process. Since five virtual cameras show an impressive improvement over the whole process, we chose five as the number of virtual cameras for our pre-training process.

\smallsec{Sampling Strategy.}
Uniform sampling makes covering the whole parameter space $\mathcal{S}$ hard.
However, our sampling strategy could cover the whole parameter space $\mathcal{S}$, thus resulting in better performance, as shown in \tabref{ab:sampling_strategy}.
Based on the observation, we use the equivalence point strategy to choose the parameters of the virtual camera.
To reduce errors, we conducted experiments with uniform sampling three times and averaged the metrics.

\begin{table}[tb]\small
  \tablestyle{3pt}{1.1}
\begin{center}
\caption{
  Ablation studies on virtual camera sampling strategy. 
  Rand represents leveraging uniform distribution as the strategy.
  The results of Rand are derived from the average of three trials to minimize errors.
}
\vspace{-2mm}
\label{ab:sampling_strategy}
\begin{tabular}{c||ccc}
\toprule
 \twRow{\textbf{Setting}} & $\times$100 & $\times$250 & $\times$300    \\
 & PSNR/SSIM  & PSNR/SSIM  & PSNR/SSIM    \\
\midrule
Rand & 41.5253/0.9489 & 39.2755/0.9283 & 36.3940/0.9070 \\
{\bf{Ours}} & \abbest{41.9842}/\abbest{0.9539} & \abbest{39.3419}/\abbest{0.9317}  & \abbest{36.6728}/\abbest{0.9147} \\
\bottomrule
\end{tabular}
\end{center}
\vspace{-2mm}
\end{table}
\setlength{\tabcolsep}{8pt}

\smallsec{Initialization of CSA for Target Camera.}
Given the initialization of CST$^T$ as described in \secref{ft_few}, we present the PSNR/SSIM difference between $\mathbf{(1,0)}$ initialization and model averaging. The results indicate that, in most scenarios, model averaging yields superior performance.
Furthermore, the performance on the Sony A7S2 of SID~\cite{chen2018learning}, as shown in \tabref{tab:csa_init}, is considered representative of the generalization ability, owing to the scale of the dataset.

\begin{figure}
    \centering
    \begin{overpic}[width=0.48\textwidth]{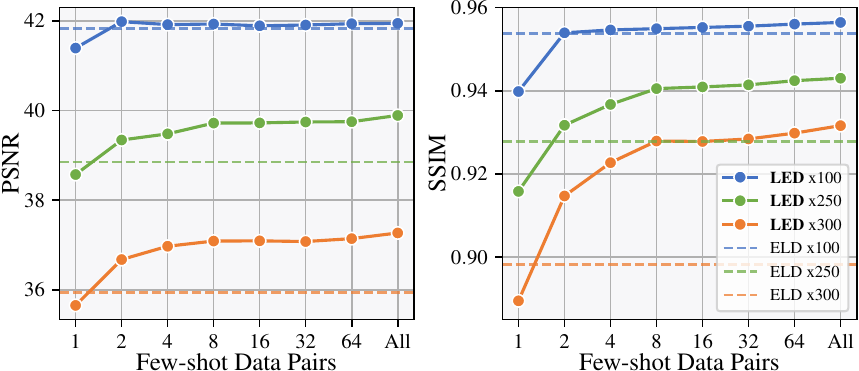}
    \end{overpic}
    \caption{
      Ablation studies on the data amount for fine-tuning that \framework~achieves superior performance with just 2 pairs for each ratio.
    }
    \label{fig:fewshot_list}
    \vspace{1mm}
\end{figure}

\smallsec{Fine-tuning with More Images.}
Ablation studies are conducted to explore the impact of the number of fine-tuning, illustrating the potential of our proposed \framework.
As depicted in \figref{fig:fewshot_list}, an increase in the quantity of paired data correlates with a gradual performance improvement. Moreover, \framework~outperforms ELD~\cite{wei2021physics} even when fine-tuning only two noise-clean pairs. Further discussions are provided in \secref{sec:discuss}.

\subsection{Further Application}
\label{sec:application_exp}

\smallsec{Equip RepNR block on other network architecture.}
By simply replacing the convolutional operators of other structures with our proposed RepNR Block, \framework~can be easily migrated to architectures beyond UNet.
In \tabref{tab:architecture}, we experimented with Restormer~\cite{zamir2022restormer} and NAFNet~\cite{chen2022simple}, transformer-based and convolution-based, respectively. 
Results demonstrate that \framework~still possesses performance comparable to calibration-based methods.

\begin{table}[tb]
  \tablestyle{2pt}{1.1}
\begin{center}
\caption{
  Experiments on network architecture. For \framework, we first replace most of the convolution block into our proposed RepNR block during pre-training and fine-tuning in deploying, \framework~outputs the same architecture as other methods without any additional computational burden, owing to the structural reparameterization procedure.
}
\label{tab:architecture}
\begin{tabular}{ccccc}
\toprule
\twRow{Architecture} &\twRow{Method}  & $\times 100$    & $\times 250$    & $\times 300$    \\
                              &                         & PSNR/SSIM     & PSNR/SSIM     & PSNR/SSIM     \\\midrule
\multirow{3}{*}{{Restormer~\cite{zamir2022restormer}}}    & P-G                     & 39.457/0.8943 & 33.956/0.7525 & 30.964/0.6409 \\
                              & ELD~\cite{wei2021physics}& \abbest{42.568}/\abbest{0.9536} & 38.699/\abbest{0.9280} & 35.863/0.9059 \\
                              & \framework                     & 42.452/0.9492 & \abbest{39.376}/0.9143 & \abbest{36.322}/\abbest{0.9143} \\\midrule
\multirow{3}{*}{NAFNet~\cite{chen2022simple}}       & P-G                     & 39.388/0.8945 & 33.892/0.7541 & 30.948/0.6445 \\
                              & ELD~\cite{wei2021physics}& 42.351/0.9535 & 38.697/0.9300 & 35.931/0.9112 \\
                              & \framework                     & \abbest{42.368}/\abbest{0.9532} & \abbest{39.277}/\abbest{0.9351} & \abbest{36.292}/\abbest{0.9188} \\\bottomrule
\end{tabular}
\vspace{-2mm}
\end{center}
\end{table}

\begin{figure}
  \centering
  \begin{overpic}[width=0.48\textwidth]{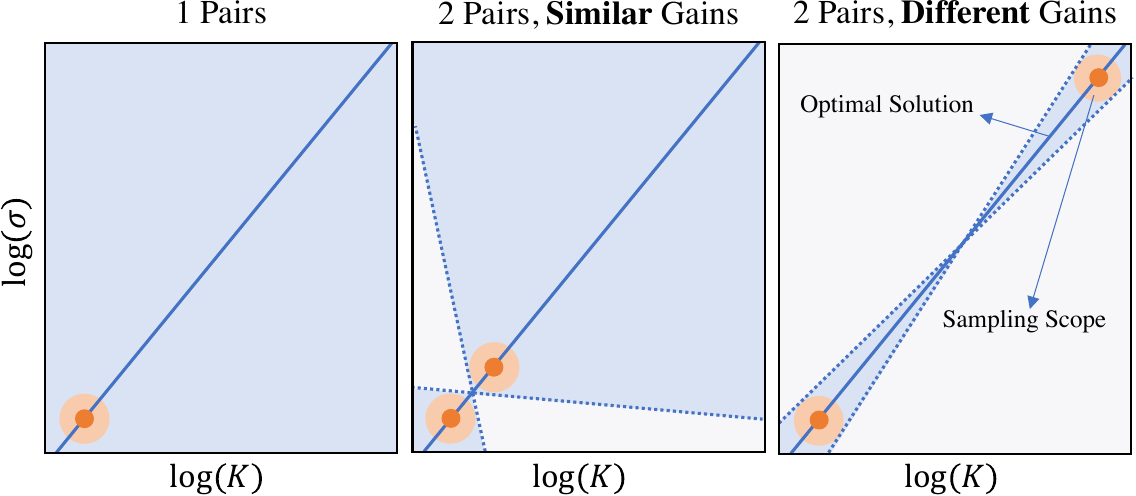}
  \end{overpic}
  \caption{
    Illustration of the feasible solution space ({\color{solution_area}{blue}} area) depicting the linear relationship between the overall system gain $\log(K)$ and noise variance $\log(\sigma)$ under various sample strategies.
  }
  \label{fig:two_pairs}
\end{figure}

\smallsec{\framework~pre-training could boost the performance of other methods.}
By integrating \framework~pre-training into various existing calibration-based or paired data-based methods, as referenced in~\cite{wei2021physics,chen2018learning}, our approach facilitates notable enhancements in performance as shown in \tabref{tab:boost}. 
These improvements are not uniform but rather depend on the difference in the pre-training strategies employed. This proves particularly effective in industrial applications, where the demands for efficiency are paramount. The strategic application of \framework~pre-training not only boosts the performance of the denoiser but also paves the way for more advanced, adaptable, and efficient denoising.

\begin{table}[tb]
  \tablestyle{4pt}{1.1}
\begin{center}
\caption{
  Experiments on \framework~pre-training with other methods. {\footnotesize{$\mathbf{X}+\mathbf{Y}$}} denotes {\footnotesize{$\mathbf{X}$}} method is training on the pre-trained network of {\footnotesize{$\mathbf{Y}$}}. {\footnotesize{$\star$}} indicates the utilization of the advanced training strategy same as PMN~\cite{feng2022learnability} for the denoiser during pre-training.
}
\label{tab:boost}
\begin{tabular}{ccccccc}
\toprule
\twRow{Method} & \twCol{$\times 100$}  & \twCol{$\times 250$}  & \twCol{$\times 300$}  \\
                        & PSNR&SSIM     & PSNR&SSIM     & PSNR&SSIM     \\\midrule
ELD~\cite{wei2021physics}                     & 41.827&0.9538 & 38.849&0.9278 & 35.940&0.8982 \\
ELD~\cite{wei2021physics}$+$\framework                 & 42.170&0.9558 & 39.285&0.9302 & 36.384&0.9058 \\
ELD~\cite{wei2021physics}$+$\framework$\star$          & \abbest{42.471}&\abbest{0.9567} & \abbest{39.454}&\abbest{0.9333} & \abbest{36.534}&\abbest{0.9138} \\\midrule
SID~\cite{chen2018learning}                     & 41.727&0.9531 & 39.135&0.9304 & 37.363&0.9341 \\
SID~\cite{chen2018learning}$+$\framework                 & 42.277&0.9580 & 39.576&0.9445 & 37.518&\abbest{0.9369} \\
SID~\cite{chen2018learning}$+$\framework$\star$          & \abbest{42.320}&\abbest{0.9585} & \abbest{39.613}&\abbest{0.9455} & \abbest{37.614}&\abbest{0.9369} \\
\bottomrule
\end{tabular}
\end{center}
\vspace{-1mm}
\end{table}

\begin{table}[t]\small
  \tablestyle{3pt}{1}
\begin{center}
\caption{
  Ablation studies on the initialization strategy of CSA for the target camera. ``Sony A7S2\#'' denotes that fine-tuning and testing is performed on the SID~\cite{chen2018learning} dataset, while other evaluations are conducted based on the ELD~\cite{wei2021physics} dataset.
}
\label{tab:csa_init}
\vspace{0mm}
\begin{tabular}{ccccccc}
\toprule
\twRow{Init}          & \twRow{Metric} & \twCol{Sony}                              & \multicolumn{1}{c}{Nikon} & \twCol{Canon}                                \\ \cmidrule{3-7} 
                               &                         & \multicolumn{1}{c}{\footnotesize{A7S2\#}} & \multicolumn{1}{c}{\footnotesize{A7S2}} & \multicolumn{1}{c}{\footnotesize{D850}}  & \multicolumn{1}{c}{\footnotesize{EOS700D}} & \multicolumn{1}{c}{\footnotesize{EOS70D}} \\ \midrule
\twRow{$\mathbf{(1,0)}$} & PSNR                    & 39.015                  & 47.310                    & 45.790                   & 41.409                     & 42.344                    \\
                               & SSIM                    & 0.9307                   & 0.9809                     & 0.9737                    & 0.9408                      & 0.9520                     \\ \midrule
\twRow{Avg.}          & PSNR                    & \abbest{39.161}                  & \abbest{47.616}                    & \abbest{45.903}                   & \abbest{41.516}                     & \abbest{42.495}                    \\
                               & SSIM                    & \abbest{0.9322}                  & \abbest{0.9817}                     & \abbest{0.9743}                    & \abbest{0.9412}                      & \abbest{0.9524}                     \\ \bottomrule
\end{tabular}
\end{center}
\end{table}

\begin{table}[tb]\footnotesize
  \tablestyle{3pt}{1}
\begin{center}
\caption{
  Ablation studies on the pair count for fine-tuning testing on the synthetic dataset.
  $n$ represents fine-tuning $n$ data pairs with a similar overall system gain for each ratio.
  $n^*$ denotes pairs of data with marginally different overall system gains.
}
\label{tab:pairs}
\begin{tabular}{ccccc}
\toprule
Ratio        & 1             & 2             & 4             & 2*                              \\\midrule
$\times 100$ & 41.295/0.9480 & 41.704/0.9523 & 41.432/0.9466 & \abbest{43.795}/\abbest{0.9648} \\
$\times 250$ & 39.239/0.9350 & 39.410/0.9351 & 39.327/0.9367 & \abbest{41.311}/\abbest{0.9457} \\
$\times 300$ & 38.314/0.9229 & 38.486/0.9216 & 38.499/0.9240 & \abbest{39.190}/\abbest{0.9278} \\\bottomrule
\end{tabular}
\end{center}
\vspace{-2mm}
\end{table}
\setlength{\tabcolsep}{8pt}

\section{Discussions}
\label{sec:discuss}

\smallsec{Why $2$ pairs for each ratio?}
As indicated in~\eqnref{eq:2pair}, the variance of noise $\log(\sigma)$ exhibits a linear relationship with the overall system gain $\log(K)$. With only one pair of data, establishing the correct linear relationship is unattainable, resulting in suboptimal performance, as demonstrated in \tabref{tab:pairs}. Furthermore, utilizing two or more pairs with similar system gains fails to precisely model the linear relationship due to a non-negligible error in the sampling scope ($\hat{\sigma}$ in \eqnref{eq:2pair}), as illustrated in~\figref{fig:two_pairs}. Following the principle of using two points to determine a straight line, adopting two pairs with marginally different system gains facilitates the accurate modeling of linearity, significantly enhancing denoising capabilities. Additionally, as shown in Fig.~\ref{fig:fewshot_list}, an increase in the number of pairs enables a more accurate fitting of linearity, thereby reducing regression errors further.

For typical explicit calibration-based methods, the primary objective of the calibration process is to compute the linear relationships mentioned previously. Subsequently, the network is trained on synthetic data to learn this relationship. However, our implicit calibration adjusts the learned linear relationships of the network directly through ``calibrating'' network parameters. This approach makes the entire process more direct and enables the network to serve as a swift adapter.

\begin{figure}
    \centering
    \begin{overpic}[width=0.48\textwidth]{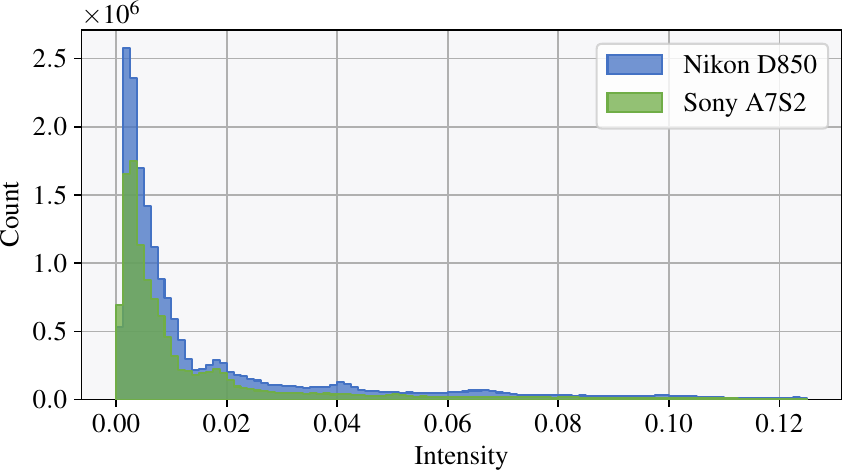}
      \put(69.3, 19){\small{$KLD=0.0289$}}
    \end{overpic}\\\vspace{2mm}
    \begin{overpic}[width=0.48\textwidth]{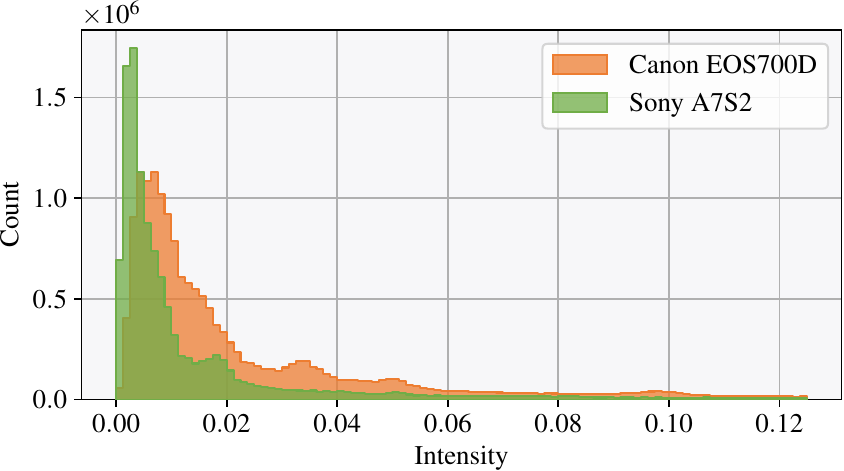}
      \put(69.3, 22){\small{$KLD=0.2978$}}
    \end{overpic}
    % \vspace{3mm}
    \caption{
        Histogram of intensities captured in the same scene with three different camera models: Nikon D850, Canon EOS700D, and Sony A7S2.
        $KLD$ denotes the KL divergence between distributions.
        Note that the distribution is similar between Nikon and Sony, 
        while the difference remains between Sony and Canon.
    }
    \label{fig:hist_src}
\end{figure}

\smallsec{Noise prior or image prior? Both!} 
It is well known that existing calibration-based methods uniformly utilize noise prior techniques (explicit noise model calibration).
However, these methods can exhibit sudden performance degradation on certain cameras, as shown in Canon EOS70D and Canon EOS700D of \tabref{tab:eld_comp},
This is attributed to these methods having learned an excessive amount of image priors from other cameras during training.
Sensors of various manufacturers would hold diverse response models,
thus yielding different signal intensities to the same scenario.
In most calibration-based methods~\cite{wei2021physics,zhang2021rethinking}, the network's denoising ability is restricted to a certain image distribution prior, \ie Sony A7S2.
As stated in~\cite{prabhakar2021few} and shown in \figref{fig:hist_src}, 
the intensity distributions of Nikon D850 and Sony A7S2 show high similarity.
Therefore, generated from the response intensity of Sony A7S2 and the noise model of Nikon D850, 
the synthetic image exhibits slight discrepancy from the real image prior,
assisting network to achieve great performance, as shown in Nikon D850 of \tabref{tab:eld_comp} of the main paper.
On the contrary, the intensity distributions between Canon EOS700D and Sony A7S2 remain large discrepancy, 
leading to a performance drop.

However, it is important to note that as additional digital gain increases, the performance gap between \framework~and other methods is gradually narrowing. This is because higher digital gain leads to more pronounced noise, making the noise prior to learning by the network more effective. Conversely, under conditions of low digital gain, the image prior previously learned by the network becomes predominant.

Based on this observation, the balance between image prior and noise prior is the key to this problem.
With the help of the proposed CSA, features are aligned to the shared space before denoising,
decreasing the influence of the image prior to the network. 
As shown in~\tabref{tab:sony_canon}, even pre-trained with the response model of Sony A7S2, \framework~can outperforms other calibration-based methods.
Furthermore, fine-tuning a few pairs of images of the target camera complements the camera-specific information, supporting the network to step forward for learning both image prior and noise prior.

\begin{table}[t]\small
  \tablestyle{.9pt}{1.1}
\begin{center}
\caption{
  Ablation studies on training with noisy pairs generated from different RAW sources. 
  The experiments are based on the Canon EOS700D camera and Sony A7S2 of the ELD~\cite{wei2021physics} dataset.
  {\bf{RAW Src.}} denotes that the RAW image pairs for fine-tuning are generated by the ground truth of Sony A7S2 or Canon EOS700D.
}
\label{tab:sony_canon}
\begin{tabular}{c||cccc}
\toprule
 \twRow{\textbf{RAW Src.}} & $\times$1 & $\times$10 & $\times$100 & $\times$200    \\
 & PSNR/SSIM  & PSNR/SSIM  & PSNR/SSIM & PSNR/SSIM    \\
\midrule
Sony  & 44.27/\abbest{0.992} & 42.15/0.982 & 37.43/0.917 & 34.74/0.867 \\
Canon & \abbest{46.24}/\abbest{0.992} & \abbest{44.14}/\abbest{0.983} & \abbest{37.94}/\abbest{0.920} & \abbest{34.78}/\abbest{0.869} \\
\bottomrule
\end{tabular}
\end{center}
\end{table}
\setlength{\tabcolsep}{8pt}

\section{Conclusion and Future Work}
\label{sec:conclusion}

To address the inherent shortcomings of calibration-based methods, we introduce a implicit calibration pipeline designed to lighting even the darkest scenes. Leveraging the camera-specific alignment (CSA), we substitute the explicit calibration procedure with an implicit learning process on the denoiser. The CSA facilitates rapid adaptation to the target camera by separating camera-specific information from the common knowledge of the noise model. Additionally, a parallel convolution mechanism is implemented to learn and eliminate out-of-model noise.
With 2 pairs for each ratio (a total of 6 pairs) and 1.5K iterations, our approach attains superior performance compared to existing methods.

Up to this point, the final output quality of \framework~is still strongly correlated with the data quality used in the few-shot fine-tuning. However, this is not solely a limitation of our method but a common drawback of most few-shot methods. Future work could focus more on making few-shot learning more stable. 
This represents a key distinction between \framework~and previous methods: earlier approaches primarily concentrated on engineering for sensor noise modeling rather than focusing on deep learning techniques like few-shot, transfer, or continual learning. Consequently, \framework~allows researchers to shift their focus from sensor engineering to exploring few-shot learning.

\section*{Acknowledgement}

This research was supported by the NSFC (NO. 62225604, 62306153) and
the Fundamental Research Funds for the Central Universities 
(Nankai University, 070-63233089).
The Supercomputing Center of Nankai University supports computation.
Moreover, we would like to express our profound gratitude to Yixuan Huang, Yipeng Du, Bowen Yin, Yunheng Li, and Ruihong Cen (in no particular order) for their dedicated efforts in constructing our dataset.

\bibliographystyle{IEEEtran}
\bibliography{main}

\newcommand{\AddPhoto}[1]{{\includegraphics[width=1in,keepaspectratio]{Authors/#1}}}
\newcommand{\AuthorBio}[3]{\vspace{-.4in}\begin{IEEEbiography}[\AddPhoto{#1}]{#2}#3\end{IEEEbiography}}

\AuthorBio{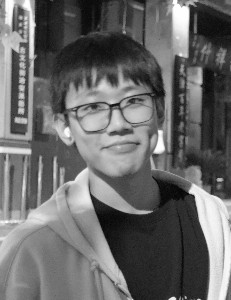}{Xin Jin}{
	received the BS degree from the College of Software, Nankai University, China, in 2022. He is currently a Ph.D. student at the College of Computer Science, Nankai University. His research interests include computational photography and video/image processing. 
  % He was a co-organizer of the MIPI workshop at CVPR 2024. 
}

\AuthorBio{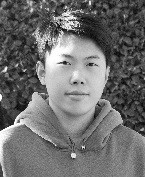}{Jia-Wen Xiao}{
  received his BS degree from the College of Computer Science, Nankai University, China, in 2022. He is currently a Ph.D. student at the College of Computer Science, Nankai University. His research interests include continual learning, self-supervised learning, few-shot learning, and computational photography.
}

\AuthorBio{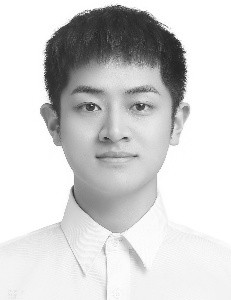}{Ling-Hao Han}{
	is a Ph.D. student from the College of Computer Science at Nankai University, under Prof. Ming-Ming Cheng’s supervision. Before that, he received a Bachelor's Degree from Nankai University in 2020. His research interests include image restoration, low-light image enhancement, and computational photography.
}

\AuthorBio{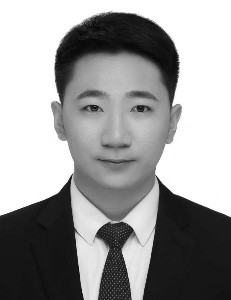}{Chunle Guo}  {
	received his PhD from Tianjin University in China. 
  He continued his research as a Research Associate with the Department of Computer Science, City University of Hong Kong (CityU), from 2018 to 2019. 
  Now, he is a postdoc research fellow working with Prof. Ming-Ming Cheng at Nankai University. 
  His research interests lie in image processing, computer vision, and deep learning.
}

\AuthorBio{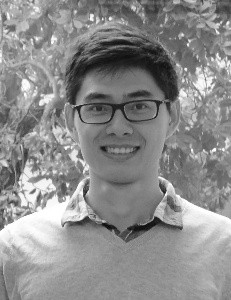}{Xialei Liu}  {
  is currently an associate professor at Nankai University, Tianjin, China. 
  Before that, he was a postdoc research associate at the University of Edinburgh, Edinburgh, UK. 
  He obtained his PhD at the Autonomous University of Barcelona, Barcelona, Spain. 
  He received B.S. and M.S. degrees at Northwestern Polytechnical University in 2013 and 2016, respectively, in Xi'an, China. 
  His research interests include continual learning, self-supervised learning, few-shot learning etc.
}

\AuthorBio{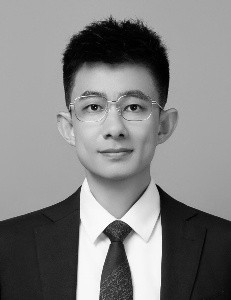}{Chongyi Li} {
  is a professor at the Nankai University, China. 
  He was a Research Fellow and then a Research Assistant Professor with City University of Hong Kong and Nanyang Technological University from 2018 to 2023. 
  His research interests include image enhancement and restoration, image generation and editing, and underwater imaging. 
  He serves as an AE of the  IEEE TCSVT, and a Lead Guest AE of IJCV. 
  %He also co-organized the MIPI workshop at ECCV 2022 and CVPR 2023. 
  He is an IEEE Senior Member.
}

\AuthorBio{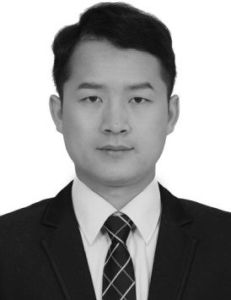}{Ming-Ming Cheng}{
	received his PhD degree from Tsinghua University in 2012,
	and then worked with Prof. Philip Torr in Oxford for 2 years.
	Since 2016, he is a full professor at Nankai University, leading the
	Media Computing Lab.
	His research interests include computer vision and computer graphics.
	He received awards, including the ACM China Rising Star Award,
	IBM Global SUR Award, \etc.
	He is a senior member of the IEEE and on the editorial boards of
	IEEE TPAMI and IEEE TIP.
}

\vfill

% that's all folks
\end{document}